\documentclass{article}

\usepackage[final]{ewrl_2023}

\usepackage[utf8]{inputenc} 
\usepackage[T1]{fontenc}    
\usepackage{hyperref}       
\usepackage{url}            
\usepackage{booktabs}       
\usepackage{amsfonts}       
\usepackage{nicefrac}       
\usepackage{microtype}      
\usepackage{xcolor}         

\usepackage{amsmath,amsfonts,bm}

\def\eqref#1{equation~\ref{#1}}

\def\1{\bm{1}}

\DeclareMathAlphabet{\mathsfit}{\encodingdefault}{\sfdefault}{m}{sl}
\SetMathAlphabet{\mathsfit}{bold}{\encodingdefault}{\sfdefault}{bx}{n}

\DeclareMathOperator*{\argmax}{arg\,max}

\usepackage{url}
\usepackage{xcolor}
\usepackage{graphicx} 
\graphicspath{ {./figures/} }
\usepackage[export]{adjustbox}
\usepackage{amssymb}
\usepackage[ruled]{algorithm2e}
\usepackage{algorithmic}
\usepackage{caption}
\usepackage{subcaption}
\usepackage{amsmath}
\usepackage{float}
\usepackage{enumitem}
\usepackage[capitalize,nameinlink]{cleveref}
\usepackage{wrapfig}

\title{APART: Diverse Skill Discovery using 
\textbf{A}ll \textbf{P}airs with \textbf{A}scending \textbf{R}eward and Dropou\textbf{T}}

\author{%
  Hadar Schreiber Galler \\
  School of Electrical Engineering\\
  Tel Aviv University\\
  and SAIPS \\
  Tel Aviv \\
  \texttt{schreiberhadar@gmail.com} \\
  \And   
  Tom Zahavy \\
  DeepMind\\
  London\\
  \texttt{tomzahavy@deepmind.com} \\
    \And   
  Guillaume Desjardins \\
  DeepMind\\
  London\\
  \texttt{gdesjardins@google.com} \\
  \And
  Alon Cohen \\
  School of Electrical Engineering\\
  Tel Aviv University\\
  and Google Research \\
  Tel Aviv \\
  \texttt{aloncohen@google.com} \\
}

\begin{document}

\maketitle

\begin{abstract}
We study diverse skill discovery in reward-free environments \citep*{VIC}, aiming to discover all possible skills in simple grid-world environments where prior methods have struggled to succeed.
This problem is formulated as mutual training of skills using an intrinsic reward and a discriminator trained to predict a skill given its trajectory.
Our initial solution replaces the standard one-vs-all (softmax) discriminator with a one-vs-one (all pairs) discriminator, and combines it with a novel intrinsic reward function and a dropout regularization technique.
The combined approach is named APART: Diverse Skill Discovery using 
\textsc{A}ll \textsc{P}airs with \textsc{A}scending \textsc{R}eward and Dropou\textsc{T}. 
We demonstrate that APART discovers all the possible skills in grid worlds with remarkably fewer samples than previous works.
Motivated by the empirical success of APART, we further investigate an even simpler algorithm that achieves maximum skills by altering VIC, rescaling its intrinsic reward, and tuning the temperature of its softmax discriminator.
We believe our findings shed light on the crucial factors underlying success of  skill discovery algorithms in reinforcement learning.
\end{abstract}

\section{Introduction}
In recent years there is a growing demand for autonomous tools in various challenging domains, including autonomous driving, robot control, and gaming. 
Deep Reinforcement Learning (Deep RL) has emerged as an effective and efficient solution to tackle these challenges \citep{Schrittwieser_2020, openai2019solving, Vinyals2019}.
Deep RL algorithms aim to maximize a given reward function that reflects the system's objectives, yet in many complex real-world scenarios, specifying this reward explicitly is difficult. 
This difficulty can arise from numerous reasons such as lack of explicit reward, misspecification of the reward, or due to having to weight multiple rewards that signify different, possibly contradictory, goals. 
For example, in an autonomous driving scenario, determining how should we weigh collisions versus speed, or measuring risk, can be challenging. 
Other factors such as comfort, traffic rules, and aggressiveness also need to be considered.

This issue has recently sparked a surge of research interest in tackling RL without a clear reward.
Previous work has shown that, one way of overcoming reward misspecification, is to utilize a diverse set of policies \citep{kumar2020one, zahavy2022discovering}.
These policies can later be used, for example, as a starting point for learning multiple goals, in hierarchical RL, for manual evaluation, or for pure exploration.
This approach may also be viewed as a small step towards unsupervised RL, where the goal is to learn informative representations of the environment by interacting with it and reaching simple, reward-free, self driven goals.

One method for obtaining a diverse set of policies is by letting each maximize some intrinsic reward. 
Unlike the extrinsic reward, the intrinsic reward is not derived from the environment and is chosen specifically to encourage diversity. 
Our work follows \cite{VIC} in which the intrinsic reward is derived from a discriminator (classifier), used to discriminate between the different policies (skills) based on their latest visited states. 

Prior work formulates this as a stochastic process in which a skill is drawn from a probability distribution, thereafter played in the environment. 
The skills and the discriminator are trained jointly as to maximize the mutual information (MI) between two random variables: the skill index (latent variable) and its latest visited state.
Fixing the discriminator, an RL algorithm updates the skill associated with the current latent variable, attempting to maximize the MI as a reward function.
Fixing the skills, the discriminator is trained in a supervised manner to identify the correct latent variable from only its latest visited state. 

MI maximization methods were followed-up in \citet{DIAYN, achiam2018variational} in which the prior distribution over the skills is fixed, and the discriminator is fed with the entire trajectory of a skill rather than just the final state.
More recent approaches employ more complex solutions, such as adding an explicit exploration term \citep{DISDAIN}, incorporating additional structure in the sampling and discrimination of skills \citep{RelativeVIC, EDDICT} or using a complex multi-staged algorithm \citep{direct_then_diffuse, ExploreDiscoverLearn}.
However, these methods still fall short in discovering all possible skills even in simple environments such as tabular grid worlds. \citep{DISDAIN}

In our work, we tackle the issue of incomplete skill discovery by directly improving the classic MI maximization approach, and without adding any additional algorithmic complexity.
During our investigation, we identified that the primary cause of the issue is the structure of the rewards used to train the skills. 
In what follows, we describe our path to reach this conclusion, and as an intermediate result we describe our method APART that does discover all skills.
We demonstrate this experimentally.

We begin our investigation by replacing the discriminator's commonly used ``One vs All'' classification method with ``All Pairs'' classification, in which $\binom{K}{2}$ binary classifiers are trained to distinguish between each pair of classes \citep[see, e.g., ][]{foundations_ml, ShaiShalev2014}, where $K$ is the number of classes. 
This, however, raises the problem of how to construct a reward function for training the skills from the classifier output, which can be inferred in many ways including: using average vote or minimum vote. 
The former takes all pair comparisons that involve the current skill and calculates an average score, similarly to the one vs all reward. 
The latter chooses the minimum score, resulting in a reward that, intuitively, attempts to minimize the error compared to the ``closest'' skill.
We test both approaches and opt to use the minimum vote to achieve better skill discrimination, which immediately translates to improved diversity metrics.
To further enhance our method, we employ a weighted dropout perturbations over the reward. 
Since all trajectories begin at the same position, we give less importance to the beginning of each episode, where all skills are packed together and are difficult to differentiate, and focus more on later stages when skills are more separable.
This approach can be thought of as interpolating between discriminating based on the last state \citep{VIC} and discriminating based on the entire trajectory \citep{DIAYN}.
The combination of the All Pairs classifier with minimum vote reward and dropouts constitutes our method, which we name~APART.

Finally, we conduct a further investigation to better understand the key components that contribute to the success of APART.
This leads us to develop a simpler algorithm that alters the classic VIC algorithm with two modifications: rescaling of the skill intrinsic rewards, and changing the temperature of the softmax function. 
We then perform an empirical evaluation to demonstrate that this modified algorithm (tuned VIC) achieves performance comparable to APART.

\subsection{Additional Related Work}

Recent works on diverse skill discovery typically add additional algorithmic mechanisms on top of MI maximization.
\citet{DISDAIN} adds an exploration bonus to help the discriminator observe enough training examples and overcome pessimistic exploration.
\cite{direct_then_diffuse} include a randomly diffusing part which adds to local exploration around the last state. 
\cite{vic_revisited} suggest that the intrinsic reward is subject to bias and employ a Gaussian mixture model.
\citet{ExploreDiscoverLearn} is a three stage methodology: exploration by training a policy to induce a uniform distribution of latent variables over states, skill discovery by training an encoder-decoder to infer latent skills sampled from this distribution, and skill learning by maximizing MI. 
\cite{DADS} change the objective to embed learned primitives in continuous spaces.
Other MI skill discovery methods include \cite{DGPO, RelativeVIC, EDDICT}.
As previously mentioned, our work aims to learn skills by using MI objective without any additional goals or stages.

\section{Preliminaries}

\subsection{Diverse Skill Discovery}
\label{subsec:skill_discovery}
Following \cite{DIAYN}, we say that a set of skills is diverse if the skills are \textbf{distinguishable} and \textbf{separable}. 
Skills are distinguishable if they reach different states.\footnote{Note that skills are still distinguishable if they visit different states using a similar choice of actions.}
Skills are separable if there are large distances between them according to some metric (see \cref{subsec:metrics} for different choices of metrics).

After being proposed in \citet{VIC}, mutual information (MI) maximization became a common method for diverse skill discovery \citep{DIAYN, warde-farley2018unsupervised, FastVIC, RelativeVIC, DISDAIN}. 
To define the MI, we consider a stochastic process in which a latent categorical variable $z$, which represents a specific skill, is sampled from a prior distribution $p(z)$.
Like \citet{DIAYN}, we fix the prior distribution $p$ to be uniform over all latent variables $z$.
After $z$ is drawn, the corresponding skill is "played" in the Markov Decision Process and we observe the generated trajectory via a conditional policy $\pi(a|s,z)$.
Then, the aforementioned MI is defined between the random trajectory of the skill and the latent variable $z$. 
\paragraph{Notations.}
$\mathcal{S}$ denotes a visited state. 
$\mathcal{A}$ denotes the action. 
$\mathcal{Z} \sim p(z)$ denotes a latent variable on which we condition our policy. 
A policy conditioned on a fixed $\mathcal{Z}$ is called a skill. 
Without limitation of generality, $p(z)$ is a uniform distribution and $\mathcal{Z}$ is discrete.
$\mathcal{I}(\cdot;\cdot)$ is the Mutual information.
$\mathcal{H}[\cdot]$ is Shannon entropy. 

We formally define MI maximization as $\mathcal{I}(\mathcal{S};\mathcal{Z})$. 
Intuitively, maximizing the MI aims for a one-to-one correspondence between the skill and its resulting trajectory.
In addition, we would also like to minimize the information between actions and states: $\mathcal{I}(\mathcal{A};\mathcal{Z} \mid \mathcal{S})$, and maximize the entropy $\mathcal{H}(\mathcal{A}|\mathcal{S})$ to assist in exploration.
The full objective $\mathcal{F}(\theta)$ as defined in \cite{DIAYN} is in \cref{eq:diyan_lower_bound} in \cref{sec:mi-maximization}, and results in:
\begin{align}
\mathcal{F}(\theta)
\geq \mathcal{H}[\mathcal{A}|\mathcal{S}, \mathcal{Z}] + \mathbb{E}_{z \sim p(z), s \sim \pi(z)} [\log q_\sigma (z | s)-\log p(z)]
\end{align}
Indeed, replacing $p(z | s)$ with a parametric model $q_\phi(z | s)$ (the discriminator) provides a variational lower bound $G(\theta, \phi)$ on our objective $F(\theta)$ \citep{information_maximization}, where $\theta$ are the policy parameters.

In DIAYN, since an actor critic policy includes entropy over actions (for exploration purposes), the objective can be further simplified as the following, where $r_z(s,a)$ is the reward:
\begin{align}
\label{eq:diyan_reward}
r_z(s,a) = \log q_\sigma(z \mid s) - \log p(z).
\end{align}

\subsection{One-vs-All and All Pairs Classification} \label{sec:ap_classification}
\label{sec:ova_and_ap}

One of the goals of this work is to showcase the significant improvement in diversity metrics achieved by switching from a One-vs-All to an All Pairs discriminator. 
To that end, we first review One-vs-All and All Pairs classification methods and highlight their distinctive features.

Multi-class classification only gained attention in later years after binary classification \citep[e.g.,][]{kearns1994introduction, cortes1995support} became reasonably well-understood.
As a result, the common approach for solving multi-class classification problems involved reducing them to learning multiple binary classifiers \citep[][for example]{svm_ovo_3}, 
and two prevalent reduction schemes emerged: One-vs-All and All Pairs \citep{foundations_ml, ShaiShalev2014}.

One-vs-All (OvA) classification is the more widespread approach. 
It requires learning $K$ different binary classifiers, where classifier $i$ separates class $k_i$ from all other classes.
A typical implementation of OvA classification assigns $K$ different score values for each example, then selects the class with maximal score.
In deep neural networks this approach usually use a softmax activation function in the final layer.
The last layer outputs the $K$ score logits $o_1,\ldots,o_K$, and the network is trained to minimize the categorical cross-entropy loss:
\[
    J^{\mathrm{ova}} = -\sum_{i=1}^K y_i \log \hat y_i, \quad \text{where} \;\; \hat y_i^\mathrm{ova} = \frac{\exp(o_i)}{\sum_{j=1}^K \exp(o_j)},
\]
and where $y_i$ is the ground truth label.

In contrast to OvA, All Pairs (AP) classification\footnote{All Pairs is also sometimes known as One-vs-One classification or All-vs-All classification.} employs a classifier for each pair of classes, resulting in a total of $L = K(K-1)/2$ classifiers. 
A classifier associated with classes $i \neq j$ predicts whether an example belongs to either class $i$ or class $j$. 
Inference can be challenging for AP classifiers, since the $L$ classifiers may produce contradicting outputs, necessitating a combination of their outputs to determine the correct label (for example, using a majority vote). Nevertheless, this is not a concern in our work as will be elaborated in the sequel.

To apply AP classification in a deep network, we adopt the method of \citet{ONEVSONE}.
We consider each of the $L$ classifiers as a binary classifier whose output is in $\{\pm 1\}$, and the final network layer outputs $L$ outputs for each classifier, denoted $o_1,\ldots,o_L$.
The outputs are normalized to $[-1,1]$ using the hyperbolic tangent function: $\hat{y}_i^{\mathrm{ap}}=\tanh(o_i)$.

Given a labeled example, we assign each classifier $i$ its correct binary label $y^{\mathrm{ap}}_i$. 
To do so, we introduce a code matrix $M_c$ of size $K \times L$ where each cell $j,i$ contains the binary label for classifier $i$ when the correct label is $j$. 
The values in the code matrix are in $\{-1, 0, 1\}$, where 0 indicates that the classifier is indifferent to the result (i.e., ``don't care'').
An example for the code matrix for $K=5$ is presented in \cref{sec:appendix_ap}.
We represent the target label $y$ as a one hot vector. Multiplying it with the code matrix yields a vector of binary labels of size $L$, as in \cref{eq:target_vector}
We then train the network to minimize the average cross-entropy terms of all the pairs as in \cref{eq:cross_entropy}:

\begin{minipage}{.24\linewidth}
\begin{equation}
\label{eq:target_vector}
y^{\mathrm{ap}} = M_c^\top y. 
\end{equation}
\end{minipage}%
\begin{minipage}{.74\linewidth}
\begin{equation}
\label{eq:cross_entropy}
    J^{\mathrm{ap}}
    =
    -\frac{1}{L}\sum_{i=1}^{L} \bigg(\frac{1+y_i^{\mathrm{ap}}}{2} \log \frac{1+\hat{y}_i^{\mathrm{ap}}}{2}
    +
    \frac{1-y_i^{\mathrm{ap}}}{2} \log \frac{1-\hat{y}_i^{\mathrm{ap}}}{2} \bigg).
\end{equation}
\end{minipage}

While OvA classification are typically preferred in supervised learning due to their simplicity, reduced number of parameters, and ease of inference, AP classification has also been successful in various settings and was benchmarked alongside One-vs-All with similar performance.
These setting include support vector machines \citep{svm_ovo_1}, and deep neural networks \citep{ONEVSONE, classifier_ovo, ovo_deep}. 
Moreover, theoretical results \cite{daniely2012multiclass} have shown that AP classifiers are more powerful than OvA classifiers (for linear models) in the sense that they can represent more complex decision rules, but require more examples to generalize.
In an RL setting, given a simulator, one could generate as many training examples as needed, so AP classification could potentially improve performance ``for free''.
Somewhat contradictory to the aforementioned theoretical results, \citet{rifkin2004defense} notice that AP classifiers tend to converge much faster in practice. 
In what follows we provide further evidence to support this observation.

\section{APART: \textbf{A}ll \textbf{P}airs with \textbf{A}scending \textbf{R}eward and Dropou\textbf{T}}

\begin{figure}[h]
\begin{center}
    \includegraphics[width=0.65\linewidth]{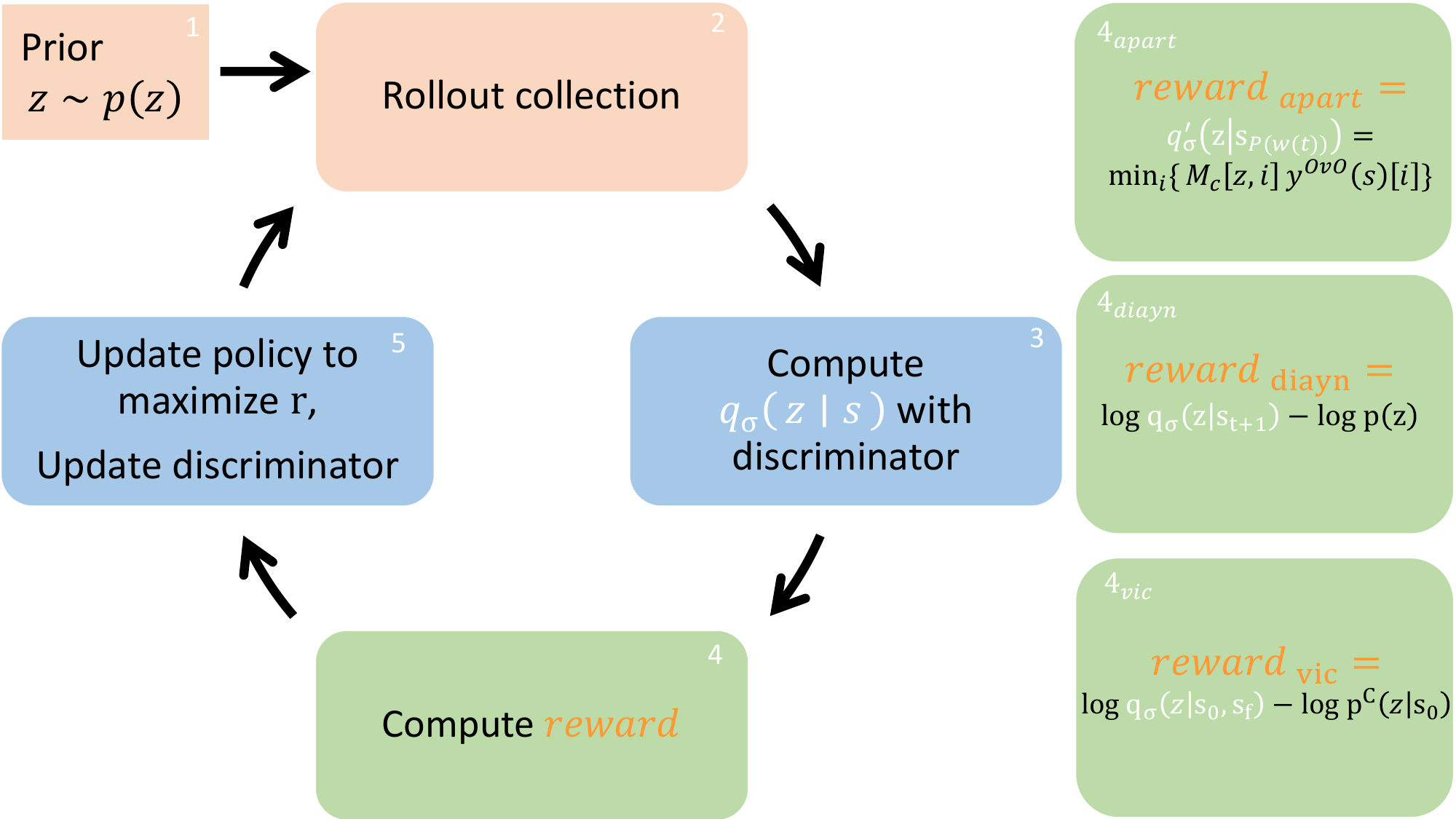}
\end{center}
\caption{The APART algorithm. Left: General scheme for mutual information maximization algorithms. (1) A latent variable $z$ is sampled from a prior distribution $p(z)$. (2) The agent interacts with the environment according to a policy $\pi(\theta,z)$. (3) Evaluate the discriminator $q_\sigma(z \mid s)$ and train it to minimize the classification loss for predicting the latent variable $z$. (4) Calculate an intrinsic reward $r_z(s,a) = \log q_\sigma(z \mid s) - \log p(z)$. (5) Train the policy to maximize $r_z(s,a)$. Right: Intrinsic rewards for DIAYN, VIC and APART.}
\label{fig:algo}
\end{figure}

In this section we introduce the APART algorithm for discovering diverse skills.
We now summarize the main elements of APART that differentiate it from prior algorithms.
These elements are outlined in \cref{fig:algo},  illustrating where each of them comes into play within APART. 
\begin{enumerate}[nosep,leftmargin=*,label=(\roman*)]
\item 
An \textbf{all pairs discriminator} (\cref{sec:ap_classification}).
\item New \textbf{intrinsic reward functions} that we derived for the AP discriminator. 
\item \textbf{Soft dropout regularization}, designed to improve the sample efficiency when using our new intrinsic reward.
\end{enumerate}

We evaluate APART experimentally in \cref{sec:experiments}, where we also ran an ablation study and demonstrate that the combination of modifications is necessary for APART's success.
In the following, we provide a detailed explanation for each of the modifications. 

\textbf{Intrinsic reward functions.}
Since the AP discriminator no longer provides the log probability of each skill, we introduce two reward functions compatible with an AP discriminator. 
To that end, recall that the standard way to classify a new example in an AP classifier is to combine all of the $L$ outputs $\hat{y}$ by multiplying them with the code matrix $M_c$ (\cref{sec:ap_classification}). 
This outputs a vector of scores for each of the $K$ classes, similarly to the output of an OvA classifier. Applying a softmax function over this vector represents a probability distribution over the classes. 

We now define our first reward function, 
called \textbf{Average All-Pairs} reward, designed to mimic the intrinsic reward of DIAYN (\cref{eq:diyan_reward}) using the output of the AP classifier.
\begin{alignat}{2}
&q_\sigma(z=y \mid s)_{average} &&= r_{average} = y_z^\mathrm{ap} \cdot \text{softmax}(M_c \hat{y}^\mathrm{ap}),
\end{alignat}
where $M_c \hat{y}^\mathrm{ap}$ are the predicted all pairs classification values multiplied with the code matrix, $y_z^\mathrm{ap}$ is a one hot vector indicating the current skill $z$ (\cref{eq:target_vector}) and $r_{average}$ is the average reward. 
In this case, $q_\sigma(z \mid s)_{average}$ can be seen as equivalent to $q_\sigma(z \mid s)$ in \cref{eq:diyan}, since it is approximating the same posterior, and therefore can be substituted to get a variational lower bound on the mutual information. 
Nevertheless, in our case we don't use the log probabilities for the reward but rather the probabilities themselves \citep[also used in ][]{RelativeVIC}.

Second, we describe the \textbf{Min All-Pairs} reward. 
This reward corresponds to taking the minimum between all pair comparisons of the current skill and all other skills \citep{zahavy2022discovering}. 
Intuitively, this leverages the structure of the AP classifier to measure distance between skills to try to separate the current skill from its ``closest'' skill.
Note that this method does not maximize the MI explicitly. See \cref{subsec:lower_bound} for an analysis on MI maximization.
Our intrinsic reward is defined as:

\begin{equation}
y_z^\mathrm{ap}(s) = [y^\mathrm{ap}_i]^T \odot \hat{y}_i^\mathrm{ap},
\quad
r_{apart} = \min_{i} \{y_z^\mathrm{ap}(s)[i]\}. \label{eq:min_ap}
\end{equation}
where $y_i^\mathrm{ap}$ is the relevant row from the code matrix (\cref{eq:target_vector}) corresponding to the latent variable, and $\hat{y}^\mathrm{ap}$ is the hyperbolic tangent applied on the prediction of the discriminator. 
The $\odot$ notation expresses element-wise multiplication of those two terms resulting in a vector size $1 \times K$. 
$y_z^\mathrm{ap}(s)$ represents pair-wise classification between target latent variable and all other latent variables, in the range $[-1, 1]$. 
Positive classification scores mean classification has correctly identified the target latent variable, while negative scores mean a misclassification (higher absolute value in the prediction is related to confidence). Therefore, when taking the minimum of this vector, we take the worst pair-wise classification score in comparison to the latent variable.

\textbf{Soft dropout regularization.}
One of the important differences between VIC and DIAYN are the time steps in which the RL algorithm receives a nonzero intrinsic reward. 
In VIC, only the terminal states are rewarded while in DIAYN all states are rewarded. 

Our first regularization technique is inspired by both of these ideas and interpolates between them. Concretely, we introduce a \textbf{time-ascending mechanism} that gives smaller weights for early time steps and larger weights for time steps that closer to the end of the episode. This property is desired when all the skills start from the same state as is usually the case since in the early time steps of the episode most skills occupy the same states, whereas in later stages they are more likely to spread out. 

We note that this mechanism is a regularization technique and we are still interested in the VIC objective, that is, to find skills that terminate at different states. More explicitly, our time-ascending weights are defined in \cref{eq:dropout}
where $t$ denotes the time step, $T$ denotes trajectory length, and $W$ is the weight. We note that \cref{eq:dropout} presents a specific choice for the weight function, but any other monotonically increasing function can be chosen instead (see \cref{fig:diff_weight_options} in the appendix for comparison with other functions).

Our second regularization mechanism uses the ascending weights as dropout probabilities to implement a random intrinsic reward. Since the expected value for a random variable $X$ for a Bernoulli distribution doesn't change, the expectation over the intrinsic reward stays the same after the dropout, implying that the optimal policy stays the same
$\mathbb{E}_{z \sim p(z), s \sim \pi(z)}[r(s,t)] = r(s).$

With these two mechanisms combined, our APART intrinsic reward is defined in \cref{eq:reward}.

\begin{minipage}{.3\linewidth}
\begin{equation}
\label{eq:dropout}
W(t)=(t/T)^2,
\end{equation}
\end{minipage}%
\begin{minipage}{.6\linewidth}
\begin{equation}
  \label{eq:reward}
  r^{\mathrm{apart}}(s,t,z)=\begin{cases}
    r(s,z), & \text{with prob. } W(t).\\
    0, & \text{otherwise}.
  \end{cases}
\end{equation}
\end{minipage}

We provide a pseudo code for the entire APART algorithm in \cref{alg:meta} in the appendix.

\section{Experiments} \label{sec:experiments}
We evaluate APART through a series of experiments on multiple grid worlds.
We first present our setup, then analyze the learned skills with comparison to prior work. Finally, we dive into each algorithm component and analyze its contribution and alternative approaches.

\textbf{Setup.}
Our goal to find all skills in simple gridworld environments, something that previous work on diverse skill discovery \citep[][and others]{DIAYN,VIC} do not achieve. 
For that, we test three different gridworld environments: four rooms, empty, and u-maze (see \cref{subsec_grid_world_envs}).
We use a single layer fully connected architecture and one hot input states (\cref{subsec_learning}). Although number of discoverable skills is different per each environment (85 in four rooms, 100 in empty and 72 in u-maze), we use a constant number of latents $N_z=100$.
We evaluate our algorithm and compare it to previous work by measuring the effective number of skills (\cref{subsec:metrics}). A more detailed setup is found in \cref{subsec_learning}

\textbf{Baselines.}
We compare our algorithm to two main baselines: DIAYN \citep{DIAYN} and VIC \citep{VIC}. 
Our version of VIC differs from the original version in that we're using a uniform prior distribution instead of learning it. 
This was shown to be a better choice in \cite{DIAYN}, and it narrows the gap between the baselines and our algorithm. 
The remaining difference, therefore, between DIAYN and VIC is the rewarded states: in VIC, only the final step is rewarded, while in DIAYN all the states are rewarded.
We also compare to DISDAIN \citep{DISDAIN} where relevant.

\subsection{Results}
We now focus on reviewing the learned skills through a series of experiments and ablation studies. 

\begin{figure}[h]
\begin{center}
  \subcaptionbox{Four Rooms\label{Rooms}}
    {\includegraphics[width=0.44\textwidth]{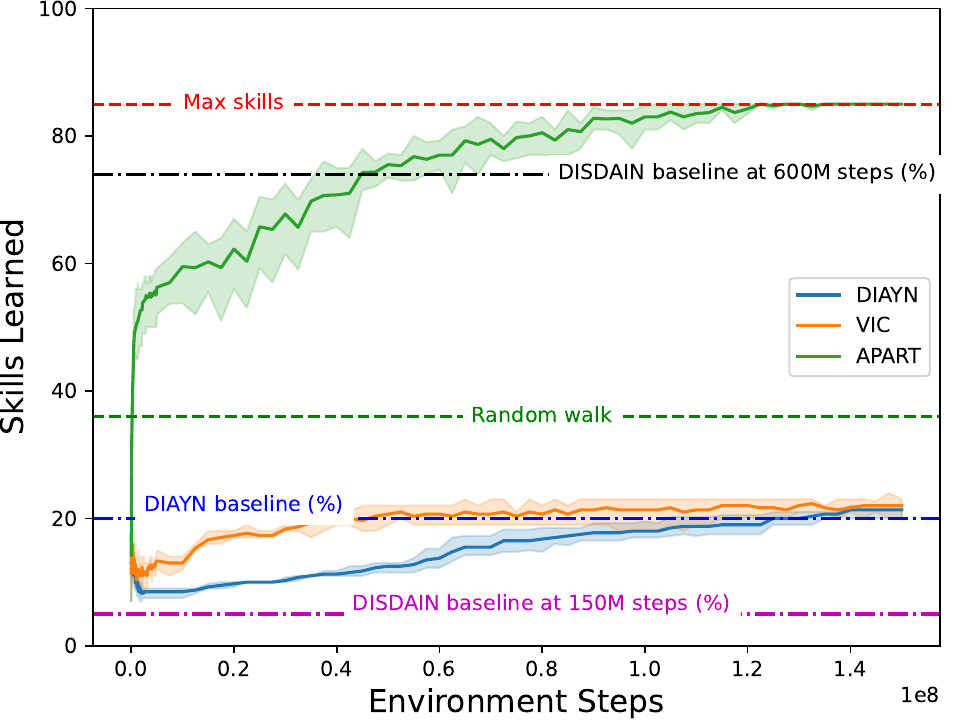}}
    \hfill
    \hfill
  \subcaptionbox{Empty\label{Empty}}
    {\includegraphics[width=0.44\textwidth]{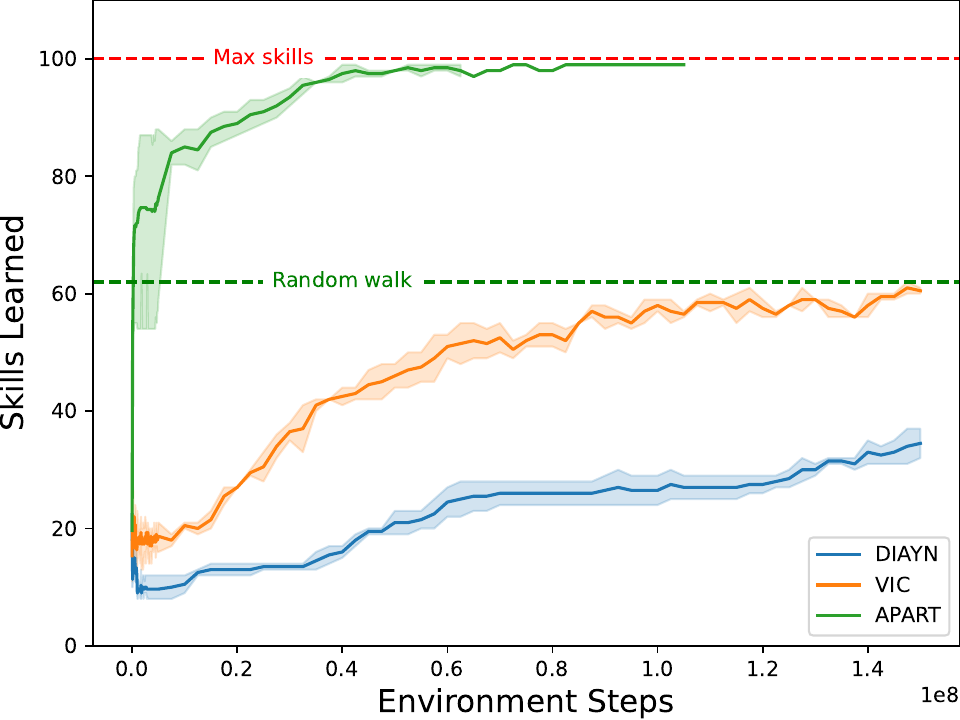}}
    \begin{minipage}[c]{0.44\textwidth}
  \subcaptionbox{U-maze\label{U-maze}}
    {\includegraphics[width=0.99\textwidth]{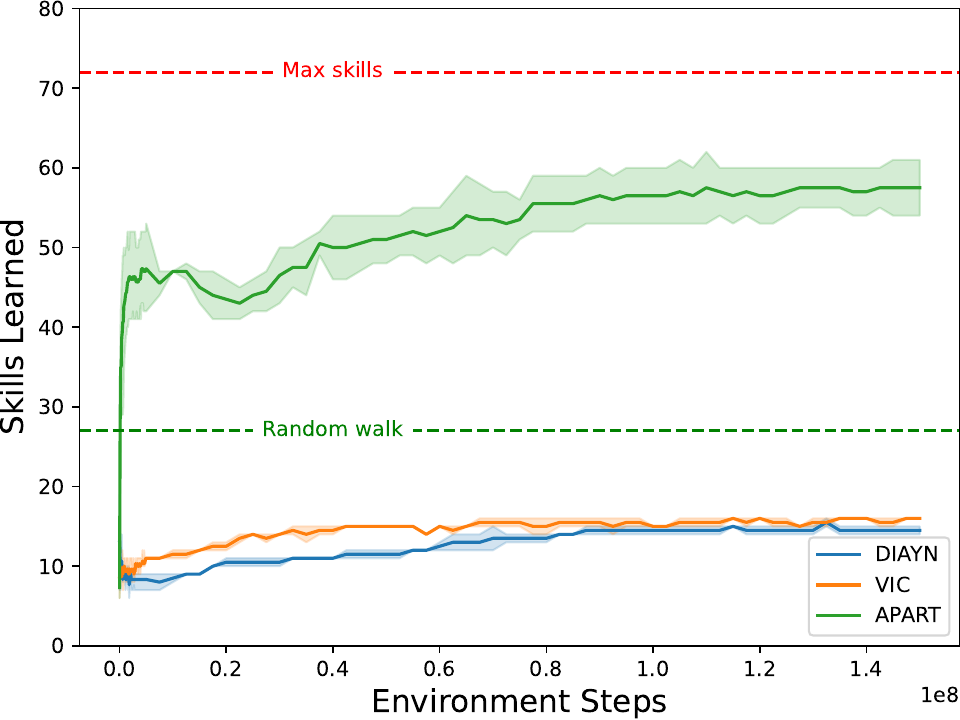}}
  \end{minipage}\hfill
  \begin{minipage}[c]{0.44\textwidth}
  \caption{\textbf{Skills learned in grid worlds:} (a) Four Rooms (b) Empty (c) U-maze. Continuous lines (mean $\pm$ std): green- APART, blue- DIAYN, orange- VIC, \textbf{dashed lines}: red- maximum available states in grid (number of unique available states in grid- effectively grid size without walls), green- mean result of a random walk agent (equally distributed actions per step), \textbf{dash-dot lines}- results reported on \citep{DISDAIN}: black- DISDAIN on 600M steps, pink- DISDAIN on 150M steps, blue- DIAYN on 40M steps.}
    \end{minipage}

\label{fig:general_learned_skills}
\end{center}
\end{figure}
\begin{figure}[h]
\begin{center}
\stepcounter{figure}
\setcounter{caption@flags}{4}

  \subcaptionbox{APART, rooms\label{fig:skills_rooms_apart}}
    {\includegraphics[width=0.32\textwidth]{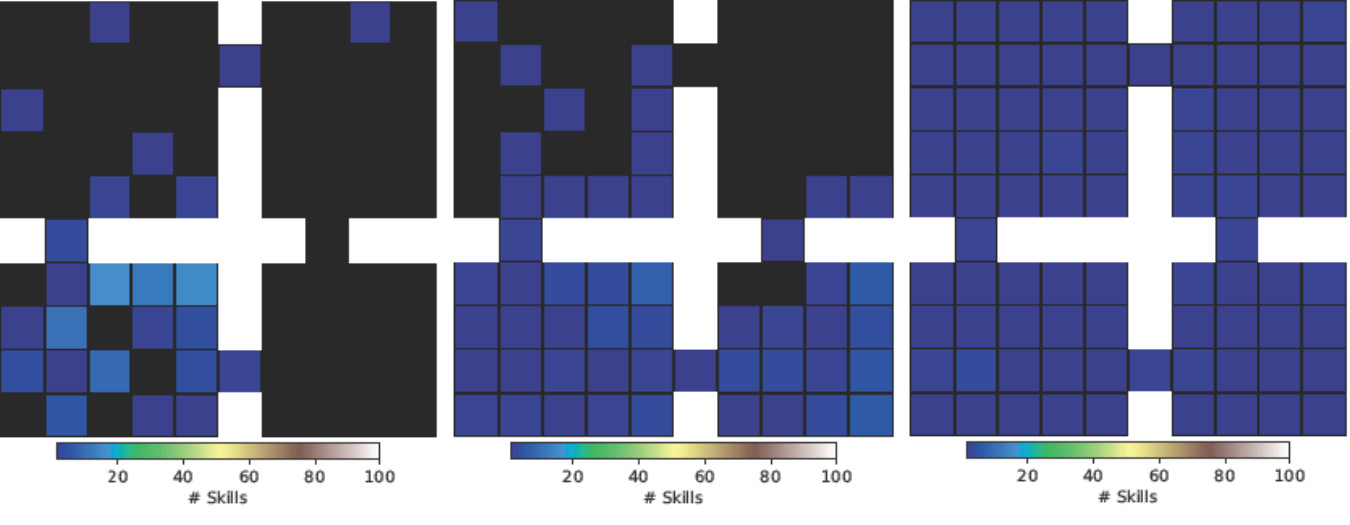}}
    \hfill
\subcaptionbox{DIAYN, rooms\label{fig:skills_rooms_diayn}}
{\includegraphics[width=0.32\textwidth]{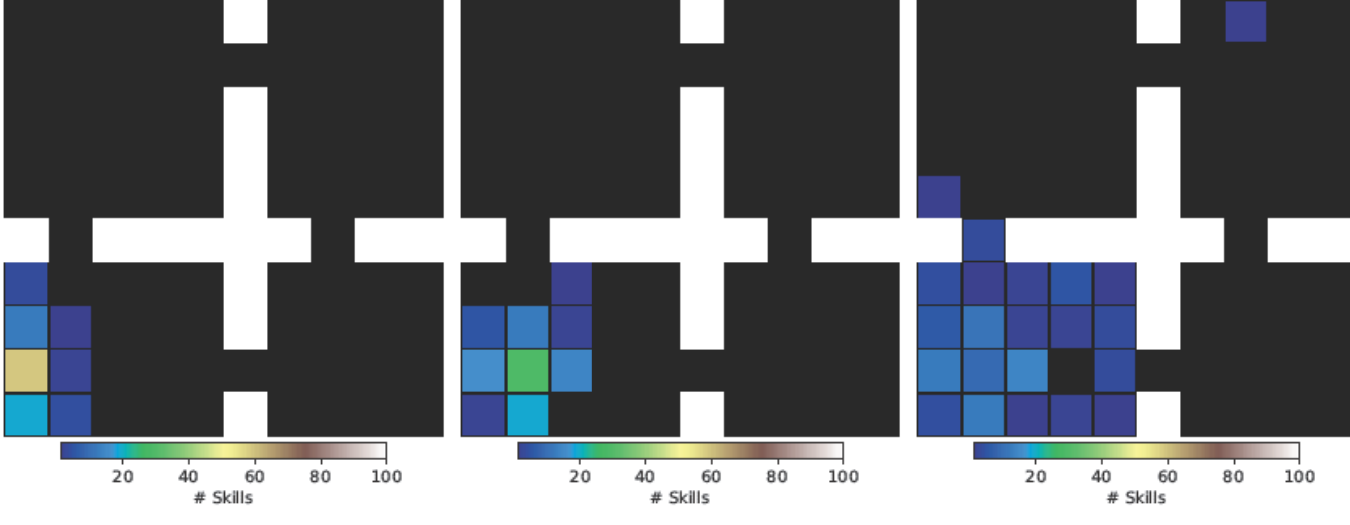}}
\hfill
  \subcaptionbox{VIC, rooms\label{fig:skills_rooms_vic}}
    {\includegraphics[width=0.32\textwidth]{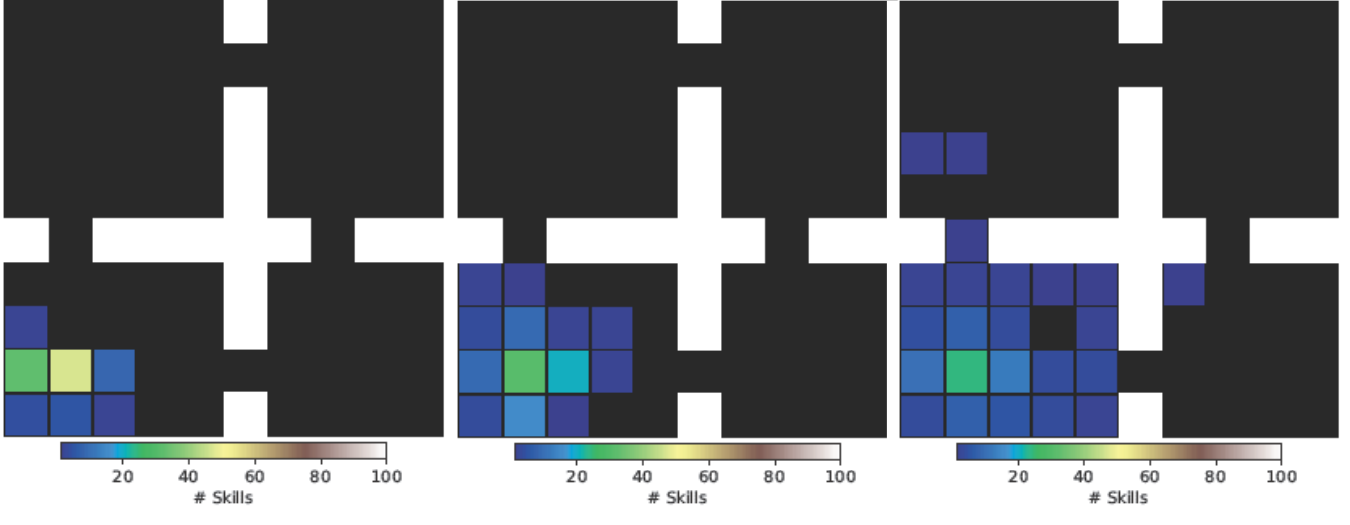}}
    \hfill
  \subcaptionbox{APART, empty\label{fig:skills_umaze_apart}}
    {\includegraphics[width=0.32\textwidth]{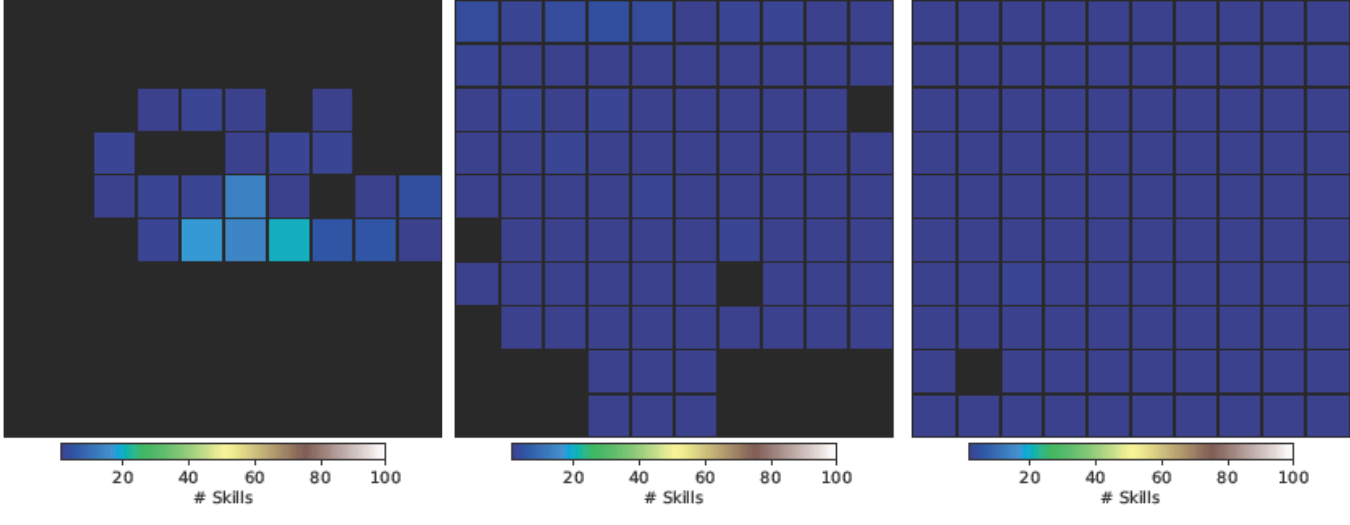}}
  \subcaptionbox{DIAYN, empty\label{fig:skills_umaze_diayn}}
    {\includegraphics[width=0.32\textwidth]{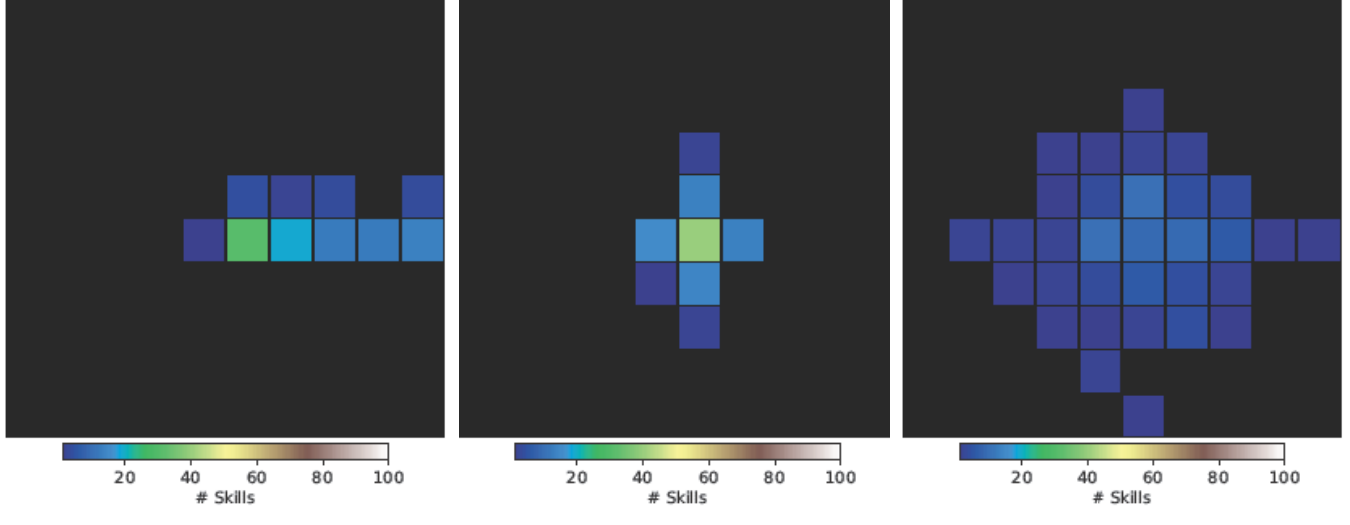}}
    \hfill
  \subcaptionbox{VIC, empty\label{fig:skills_umaze_vic}}
    {\includegraphics[width=0.32\textwidth]{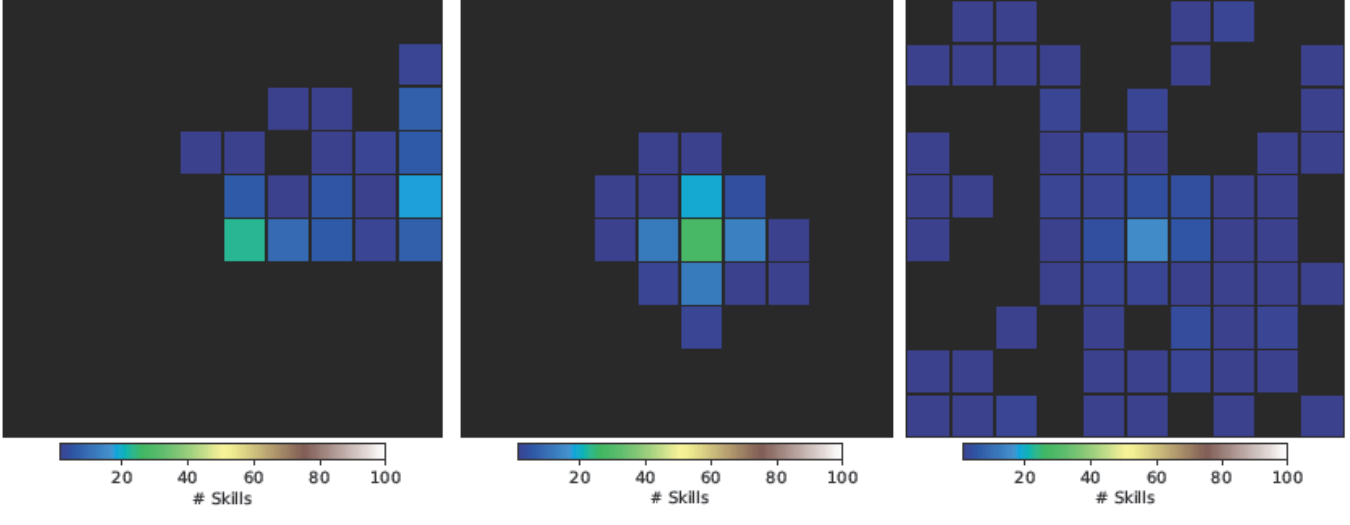}}
    \hfill
  \caption{\textbf{States reached during training} with APART (\subref{fig:skills_rooms_apart}, \subref{fig:skills_umaze_apart}), DIAYN (\subref{fig:skills_rooms_diayn}, \subref{fig:skills_umaze_diayn}) and VIC (\subref{fig:skills_rooms_vic}, \subref{fig:skills_umaze_vic}), on four rooms and empty. Each plot represents a different training point: Left to right: beginning of training, before convergence (5M steps), after convergence (150M steps).} 
\label{fig:states_reached}
\end{center}
\end{figure}
\paragraph{Number of learned skills.}
\cref{fig:general_learned_skills} provides a comparison between APART and the baselines in terms of number of learned skills. 
\subref{Rooms}: In four rooms, we see that DIAYN and VIC don't exceed 30 skills after convergence of over 100M steps \citep[similar results are reported in][]{DISDAIN}. 
DISDAIN\footnote{DISDAIN reported results are in a percentage from maximum available states (104 available states vs 85 available states in our configuration). DISDAIN uses 20 steps while we use 40} does not converge when limiting environment steps to 150M. At 600M steps, convergence achieved by the best seed is reported to be 86\% of skills.
APART achieves all skills with all seeds in approximately 100M steps with a stable convergence.
In \cref{fig:states_reached} (left side), results are supported with an example rollout before, during and after convergence. While DIAYN and VIC barely leave the first room in the four rooms environment, APART manages to not only visit all rooms but also reaches all available states.
\subref{Empty}: In empty, DIAYN and VIC converge to less than 60 skills, while APART reaches all available states after 50M steps.
In \cref{fig:states_reached} (right side), DIAYN and VIC are centered and don't reach all distant locations while APART converges to reach all available states.
\subref{U-maze}: In u-maze, DIAYN and VIC converge to less than 20 skills, while APART reaches more than 50 available skills (out of the 72 possible states).

\textbf{All Pairs vs.\ One vs All.}
Here we focus on the improvements achieved by changing our classifier to use the all pairs classification, instead of the standard one vs all method. These improvements are studied using ablations and alternative approaches.
\begin{figure}[h]
\begin{center}
  \subcaptionbox{Rooms, All States\label{rooms_all_states}}
    {\includegraphics[width=0.4\textwidth]{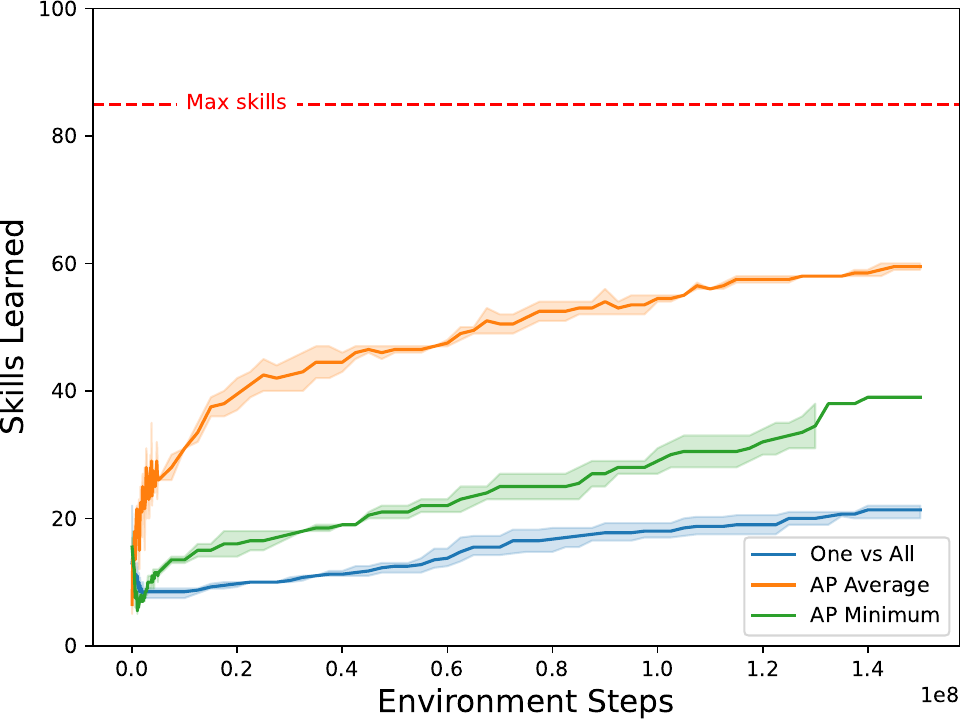}}
    \hfill
  \subcaptionbox{Rooms, Last States\label{rooms_last_states}}
    {\includegraphics[width=0.4\textwidth]{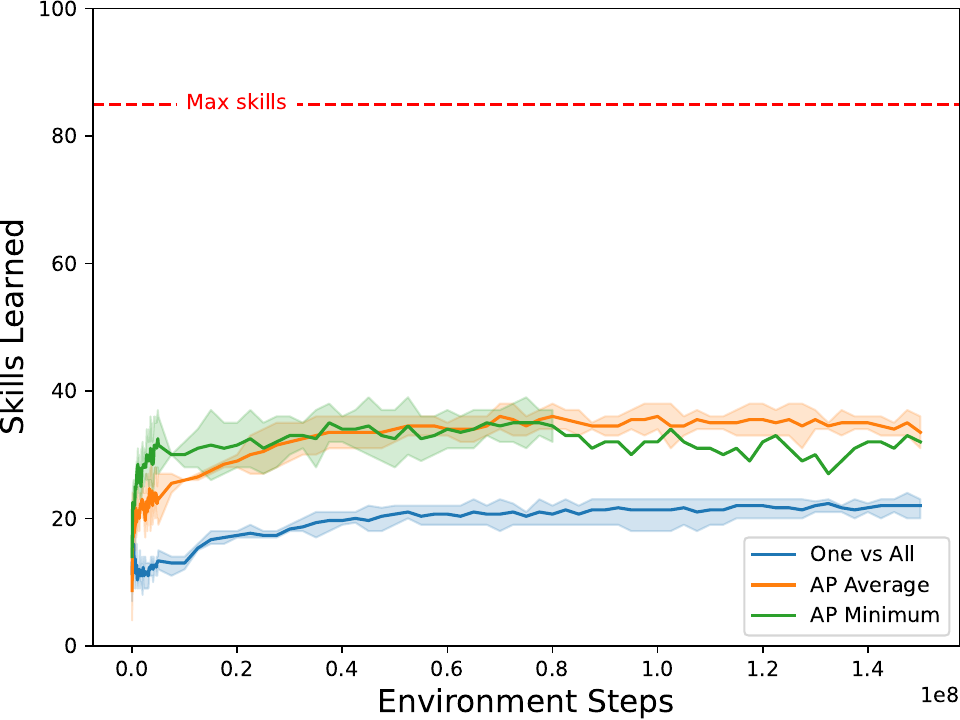}}
    \hfill
  \subcaptionbox{Empty, All States\label{empty_all_states}}
    {\includegraphics[width=0.4\textwidth]{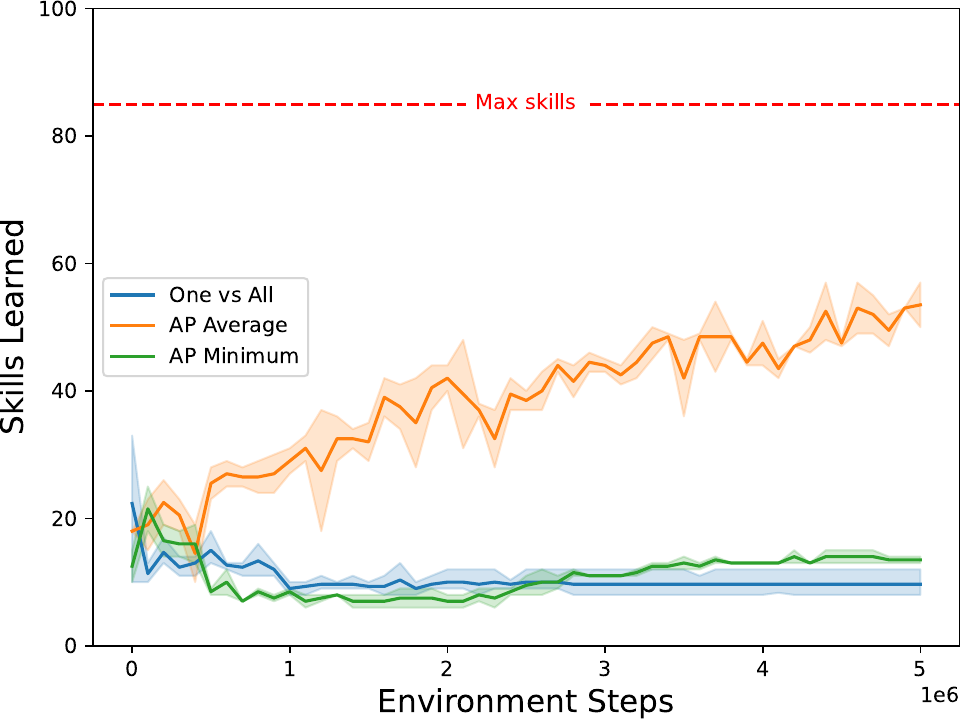}}
    \hfill
  \subcaptionbox{U-maze, All States\label{umaze_all_states}}
    {\includegraphics[width=0.4\textwidth]{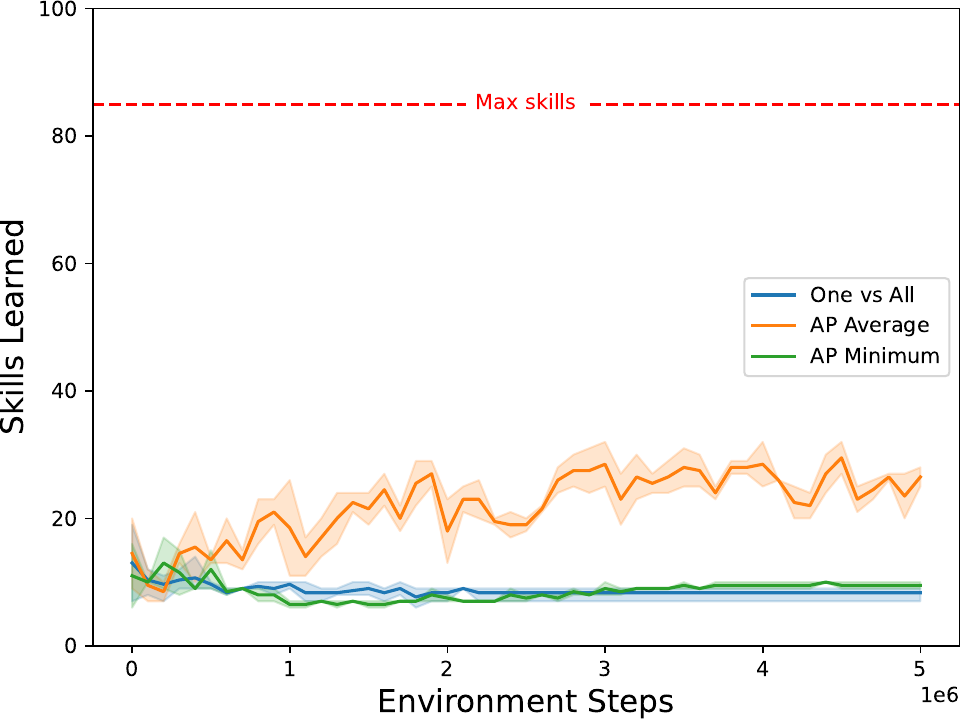}}
    \hfill
  \caption{\textbf{All pairs study.} Effect of all pairs on learning. \textbf{All States}: blue- all pairs with minimum reward, orange- all pairs with average reward, green- one vs all. \textbf{Last states}: blue- all pairs with minimum reward, orange- one vs all.}
\label{fig:all_pair_variations}
\end{center}
\end{figure}
In \cref{fig:all_pair_variations}, we see the effect of using the all pairs classifier. In \subref{rooms_all_states}, a DIAYN setup is used on rooms (all state contribute to reward), \subref{rooms_last_states} show a VIC setup on rooms (only last states contribute to reward), \subref{empty_all_states} shows using all states on empty, with 5M steps and \subref{umaze_all_states} shows using all states on u-maze with 5M steps.
Evaluations on 5M steps give us the opportunity to see the trend while saving compute resources along the way.
In all graphs, all pairs exceeds the one vs all, whether if the all pairs reward type is the average reward or the minimum reward.
When comparing the minimum reward and the average reward setups by themselves, we can see that AP Average achieves the highest scores. However, it does not achieve the maximum environment score as in \cref{fig:general_learned_skills}

\textbf{Reward type study.} 
The focus of this section is on the improvements we achieve through the addition of ascending rewards and dropouts. An ablative approach is used to study these improvements.

\begin{figure}[h]
\begin{minipage}[b]{0.49\textwidth}
    \begin{subfigure}[b]{0.5\textwidth}
    \centering
        \includegraphics[width=1\textwidth]{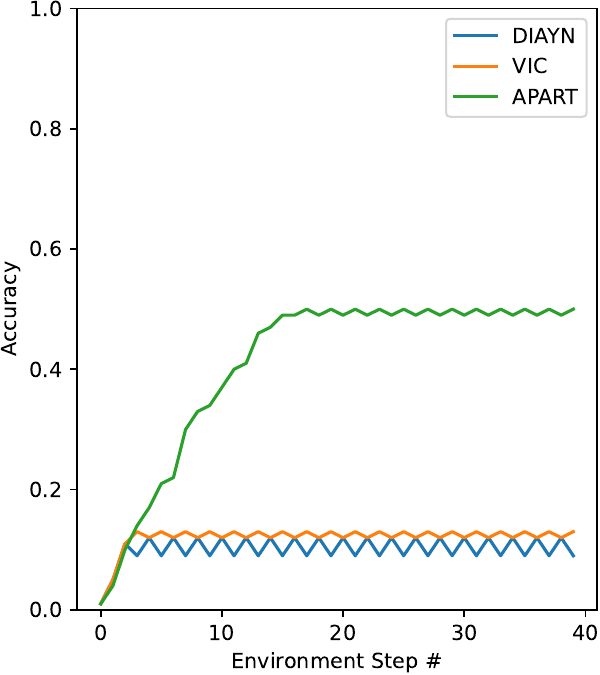}
        \caption{5M steps}
        \label{subfig:5m_env_steps}
    \end{subfigure}%
    \begin{subfigure}[b]{0.5\textwidth}
        \includegraphics[width=1\textwidth]{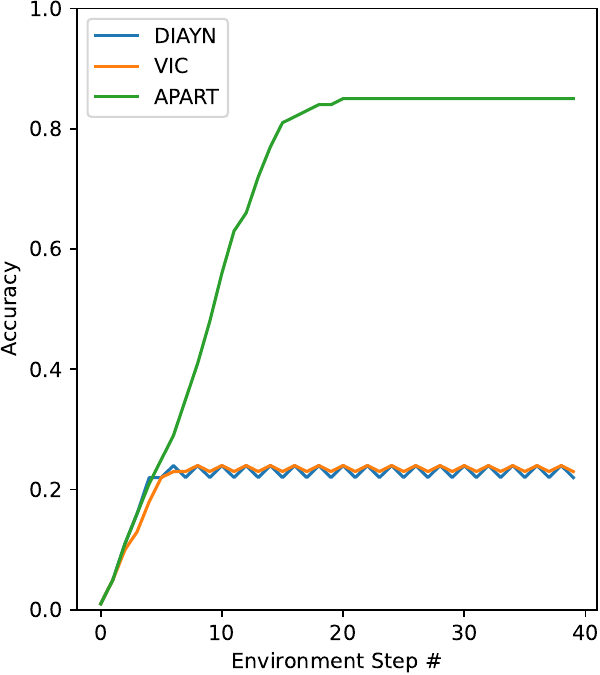}
        \caption{150M steps}
        \label{subfig:150m_env_steps}
    \end{subfigure}%
\caption{Discriminator accuracy per \\environment step, rooms.}
\label{fig:accuracy_per_step}
    \end{minipage}
    \begin{minipage}[b]{0.47\textwidth}
    \vspace{-0.5cm}
    \centering
        \hspace*{0cm}\begin{tabular}{l|l|l|l|l}
        \begin{tabular}[c]{@{}l@{}}\tiny\textbf{Classifier}\end{tabular} &
        \begin{tabular}[c]{@{}l@{}}\tiny\textbf{Reward}\end{tabular} &
        \begin{tabular}[c]{@{}l@{}}\tiny\textbf{Ascending}\end{tabular} &
        \begin{tabular}[c]{@{}l@{}}\tiny\textbf{Dropout}\end{tabular} &
        \begin{tabular}[c]{@{}l@{}}\tiny\textbf{Result,} \\ \tiny\textbf{5M steps}\end{tabular}\\  \hline
        \textbf{AP}         & Min.              & -                     & -                & 10 $\pm$ 1.5                         \\
                            & Min.              & +                     & -                & 46 $\pm$ 1                         \\
                            & Min.              & -                     & +                & 10 $\pm$ 0                         \\
                            & Min.              & +                     & +                & \textbf{56.25 $\pm$ 3.7}                         \\ \cline{2-5} 
                            & Avrg.              & -                     & -                & 26 $\pm$ 0                         \\
                            & Avrg.              & +                     & +                & 25 $\pm$ 0                        \\ \hline
        \textbf{OvA}        & Avrg.              & -                     & -                & 8.5 $\pm$ 0.8                          \\
                            & Avrg.              & +                     & -                & 12 $\pm$ 0                          \\
                            & Avrg.              & -                     & +                & 9.5 $\pm$ 2.5                         \\
                            & Avrg.              & +                     & +                & 8.5 $\pm$ 0.5                         
        \end{tabular}
        \caption{Skills Learned by Reward Type, \\ mean $\pm$ std, rooms.}
        \label{table:reward_type}
    \end{minipage}

\end{figure}

In \cref{table:reward_type} we see that while adding the ascending reward and dropout doesn't affect the average reward methods, it does affect learning with the minimum reward.
While AP is shown to always be a more successful choice, we can see that when pairing minimum reward AP with ascending reward and dropout we achieve the most learned skills.
In order to characterize the behavior during the episode, \cref{fig:accuracy_per_step} presents the discriminator accuracy per environment step $t$, in 5M steps and 150M steps, where
$\text{accuracy}=\tfrac{\text{Correct predictions}}{\text{Total predictions}}$.
The three methods achieve lower accuracy in the beginning of the episode since all skills are tightly packed. With all methods, the accuracy ascends up to a certain value and stabilizes around it.
It is important to point out that discriminator accuracy has to be paired with a policy, since it is challenging to achieve high accuracy when skills are not spread out. Nonetheless, we see that our method achieves much higher accuracy. While other methods do not exceed accuracy of 25\%, APART achieves 85\% accuracy, which is the maximum achievable accuracy since this is the number of separable states in rooms environment.

\section{Epilogue: Tuning VIC} 
\begin{wrapfigure}{h}{0.45\textwidth}
\vspace{-0.8cm}
\centering
    {\includegraphics[width=0.44\textwidth]{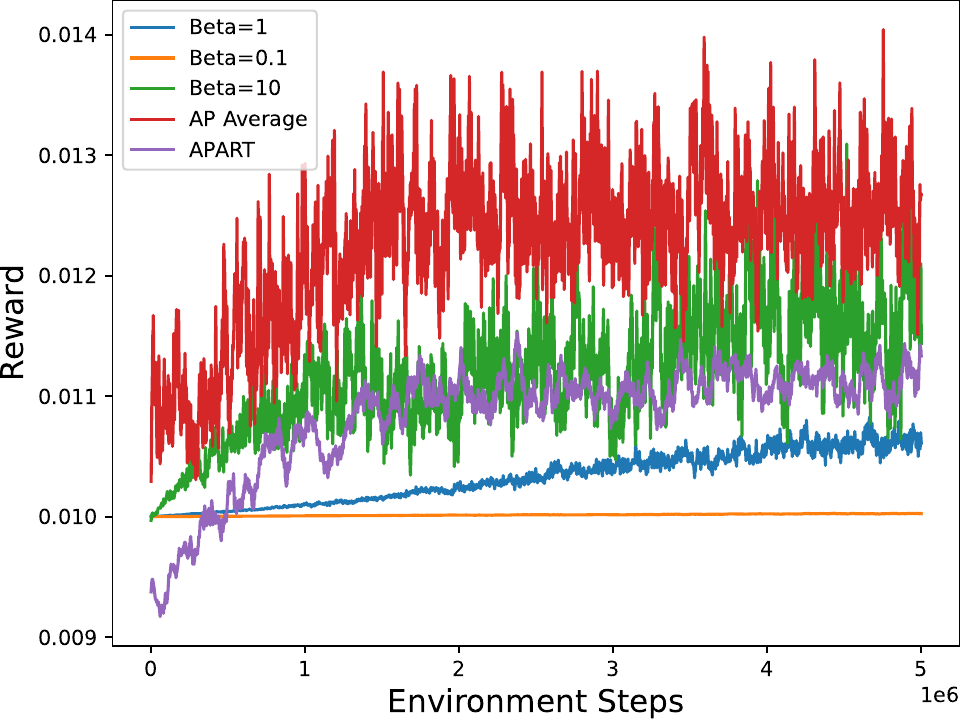}}
  \caption{Learned rewards from five different methods: APART and AP Average, Tuned VIC with: $\beta={10, 1, 0.1}$. 
  Rewards are smoothed with $\sigma=5$.}
\label{fig:reward_comparison}
\end{wrapfigure}
In this section, we modify the VIC algorithm to match performance of APART. 
It is well known that Deep RL algorithms are sensitive to the scale of the reward function, and many prior works have tried to address this challenge \citep{hessel2021muesli, kapturowski2019recurrent, hasselt2016learning}. 
Here, we take a similar approach and tune the scale of the reward of VIC to be similar to the scale of APART's reward. 
Our simple modifications allow VIC to discover all the skills in our tested grid world environments (\cref{subsec_grid_world_envs}), outperforming multiple prior works.

Our motivation follows by recalling that APART uses an AP classifier whose outputs not represent probabilities.
Instead a $tanh$ activation bounds the output in $(-1, 1)$ (\cref{sec:ap_classification}). \\
Inspired by this, we first decided to remove the  $\log$ function from the reward in VIC, an approach reported to work well in relative VIC \citep{RelativeVIC}. 
Second, we tuned the temperature of the softmax function to match the scale of APART's reward.

\cref{fig:reward_comparison} compares reward scale for three different values of $\beta$, the inverse temperature of the VIC's softmax function. 
As can be seen, for $\beta = 10$, the scale of the VIC reward matches the scale of the APART reward. 
In \cref{fig:reward_comparison_appendix} (\cref{sec:additional-results}), we also provide the scale of the vanilla VIC reward, which is observed to be larger than that of APART. 
We also note that for $\beta=10$, the probability mass is more concentrated on the outcome with the highest probability, see \cref{appendix:softmax_temp} for more details. 

\begin{figure}[h]
\begin{center}
  \subcaptionbox{Four Rooms\label{Rooms_tunedvic}}
    {\includegraphics[width=0.44\textwidth]{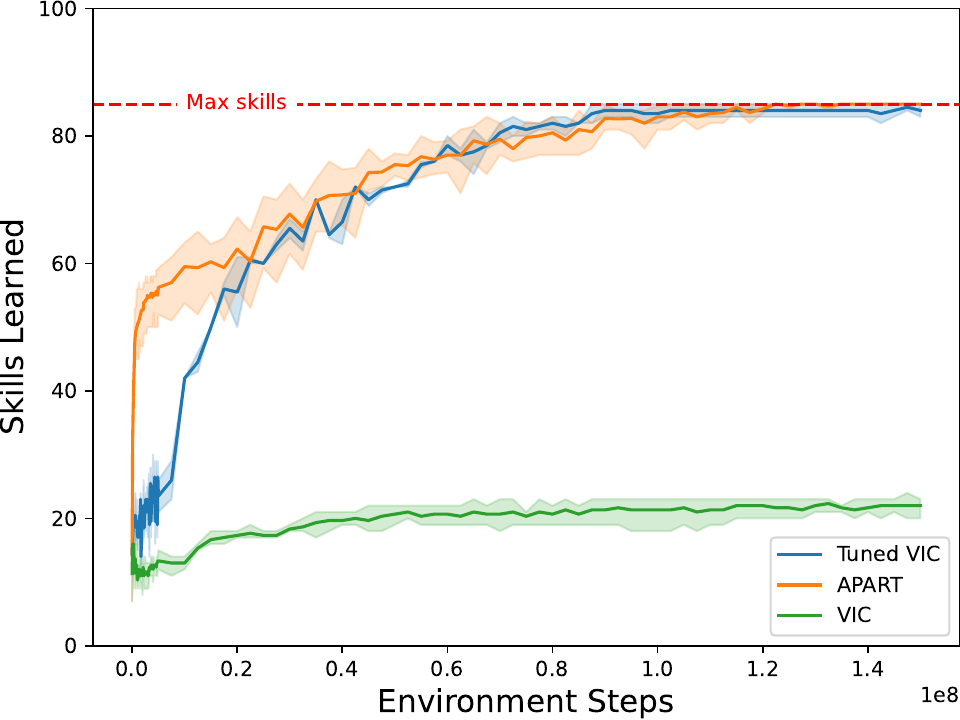}}
    \hfill
    \hfill
  \subcaptionbox{Empty\label{Empty_tunedvic}}
    {\includegraphics[width=0.44\textwidth]{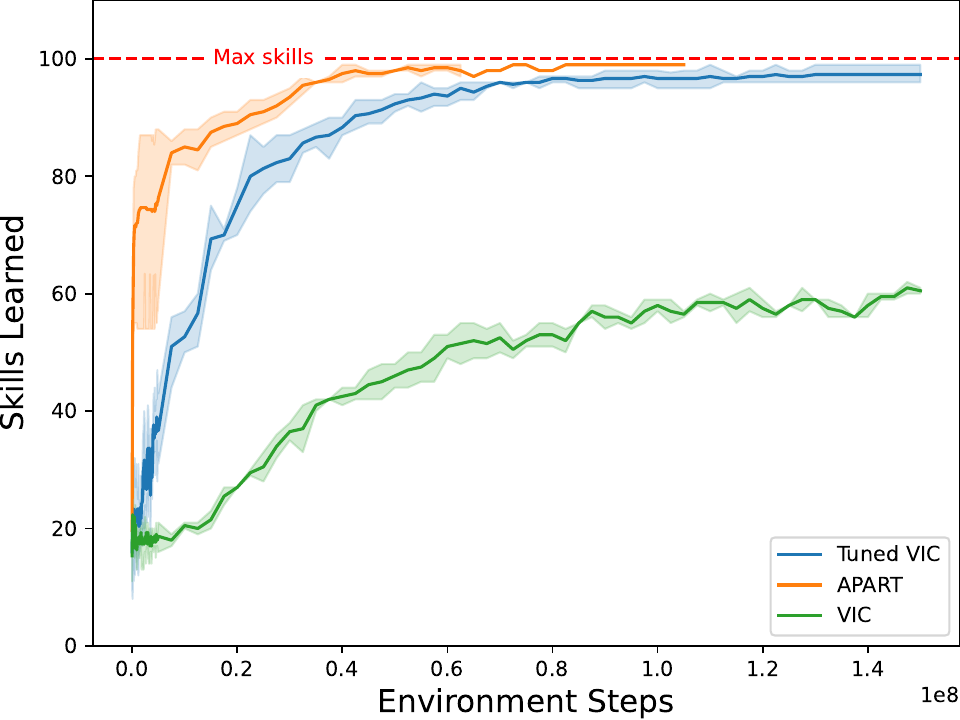}}
  \begin{minipage}[c]{0.44\textwidth}
  \subcaptionbox{U-maze\label{U-maze_tunedvic}}
    {\includegraphics[width=0.99\textwidth]{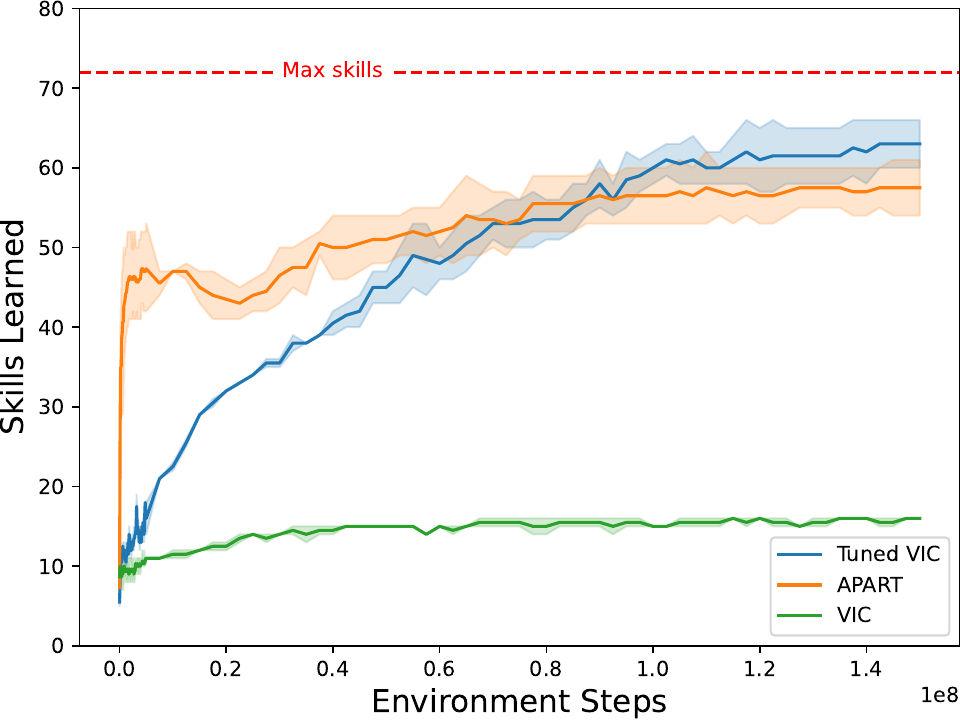}}
  \end{minipage}\hfill
\begin{minipage}[c]{0.49\textwidth}
  \caption{\textbf{Skills learned in grid worlds:} (a) Four Rooms (b) Empty (c) U-maze. Continuous lines (mean $\pm$ std): orange- APART, blue- Tuned VIC, $\beta=10$, green- VIC. \textbf{dashed red line}: maximum available states in grid (number of unique available states in grid- effectively grid size without walls).}
  \end{minipage}
\label{fig:tunedvic_learned_skills}
\end{center}
\end{figure}

\paragraph{Results.}
We evaluated the performance of the tuned VIC algorithm on the same grid world environments. 
The results are presented in \cref{fig:tunedvic_learned_skills}. 
The convergence in APART is faster, but the final performance is comparable across all three environments, with a slight advantage for tuned VIC in the U-maze and a slight advantage for APART in the empty environment.

\section{Discussion}
In this work we introduced APART, a method for skill discovery without an extrinsic reward function. 

Most previous works on diverse skill discovery are unable to discover all skills even on simple tabular environments. 
There are a few cases where tailored solutions such as adding an explicit exploration term \citep{DISDAIN} or using a complex multi-staged algorithm \citep{direct_then_diffuse, ExploreDiscoverLearn} managed to do so.
APART, on the other hand, does not change the diversity objective and does not require complicated  multi-stage training, yet it can discover all the skills in these environments. 

Furthermore, APART demonstrated superior skill discovery, compared to prior work in multiple grid world environments. 
While grid world environments are arguably a small-scale case study, our gains are significant and outperform multiple commonly used baselines.
As such we believe that our insights will be useful for future research. 
Furthermore, our ablative analysis suggests that each one of the components in APART contributed to its performance, and in particular only utilizing our discriminator achieves a much higher accuracy than those found in prior work.

Inspired from this success, we further studied the baseline VIC algorithm and showed that it can be improved to discover all the skills on grid worlds with two simple steps. These are very encouraging results, considering that the reported results on this benchmark in multiple papers \citep[][etc.]{DIAYN, DISDAIN}, were much lower.

Our results in simple grid world environments can be further investigated in more complex environments, such as ATARI \citep{atari} and Mujoco \citep{mujoco}. 
As a first step towards this, we demonstrate APART's success when using pixel observations (\cref{fig:pixel_observation}). 

Finally, we hope our study on using alternative classification techniques in diverse skill discovery will inspire others to explore more ideas in this direction, such as error correction codes and hierarchical classification. Such investigation in this domain is rather appealing since there is no limit on the amount of samples an RL agent can collect in a simulation. These may spark the definitions of richer notions of diversity. 

\section*{Acknowledgements}
AC is supported by the Israeli Science Foundation (ISF) grant no. 2250/22.

\bibliography{sample.bib}

\appendix
\section{Appendix: Background}
\subsection{Reinforcement Learning and Markov Decision Processes}
\label{sec:rl}
A discounted infinite-horizon MDP is a tuple $\mathcal{M}=(S, A, P, \gamma, r)$ where $S$ denotes the state space, $A$ denotes the action space, $P$ is the transition function, $\gamma \in [0, 1)$ is the discount factor and the reward function $r$. 
An agent interacts with the MDP as follows. At any state $s$, the agent chooses an action $a$, thereafter it receives a reward $r(s)$ and the MDP transitions to the next state, drawn from the distribution $P(\cdot \mid s,a)$.

A policy $\pi(\cdot)$ maps states to actions and defines our agent's behavior.
An optimal policy $\pi^\star(\cdot)$ is the policy that maximizes expectation over cumulative discounted reward:
$\mathbb{E}[\sum_{t=0}^\infty \gamma^t R(s_t)]$, where the expectation is taken over the state transitions, and possibly the randomness of the agent.

The action-value function associated with a policy $\pi$, also known as the $Q$-function, gives us the value (expected discounted future reward) at state $s$ with the immediate reward under action $a$, thereafter playing according to $\pi$. 
We also denote $Q^\star$ as the action-value function associated with $\pi^\star$.
$Q^\star$ has the property that $\pi^\star(s) \in \argmax_{a \in A} Q^\star(s,a)$ for all states $s$.

In reinforcement learning, we typically deal with an MDP whose reward or transition functions are unknown, and we are interested in approximating $\pi^\star$ by repeated interactions with the model.
One way of finding the optimal policy $\pi^\star$, is to estimate $Q^\star$ then choose $\pi^\star(s) = \argmax_{a \in A} Q^\star(s,a)$. 
In deep Q-Networks \cite[DQN;][]{DQN}, a deep neural network (NN) is used to estimate $Q^\star$.
The NN receives an observation space, representing the state and outputs an estimate of $Q(s,a)$, from which $Q^\star$ is chosen and acted upon.
The observation space can vary and is an implementation choice.

\subsection{All Pairs Classification}
\label{sec:appendix_ap}
An example for the code matrix for $K=5$ is:
\[
    \begin{pmatrix}
        1  & 1  & 1  & 1  & 0  & 0  & 0  & 0  & 0  & 0 \\
        -1 & 0  & 0  & 0  & 1  & 1  & 1  & 0  & 0  & 0 \\
        0  & -1 & 0  & 0  & -1 & 0  & 0  & 1  & 1  & 0 \\
        0  & 0  & -1 & 0  & 0  & -1 & 0  & -1 & 0  & 1 \\
        0  & 0  & 0  & -1 & 0  & 0  & -1 & 0  & -1 & -1
    \end{pmatrix}.
\]

\subsection{Mutual Information Maximization} \label{sec:mi-maximization}
The full objective as defined in \cite{DIAYN} is:
\begin{align}
\mathcal{F}(\theta) 
& \triangleq \mathcal{I}(\mathcal{S};\mathcal{Z}) + \mathcal{H}(\mathcal{A}|\mathcal{S}) - \mathcal{I}(\mathcal{A};\mathcal{Z}|\mathcal{S}) 
\nonumber \\
&= (\mathcal{H}[\mathcal{Z}] - \mathcal{H}[\mathcal{Z}|\mathcal{S}]) + \mathcal{H}(\mathcal{A}|\mathcal{S}) - (\mathcal{H}(\mathcal{A}|\mathcal{S}) - \mathcal{H}[\mathcal{A}|\mathcal{S}, \mathcal{Z}]) 
\nonumber \\ 
&= \mathcal{H}[\mathcal{A}|\mathcal{S}, \mathcal{Z}] - \mathcal{H}[\mathcal{Z}|\mathcal{S}] + \mathcal{H}[\mathcal{Z}] 
\nonumber \\
&= \mathcal{H}[\mathcal{A}|\mathcal{S}, \mathcal{Z}] + \mathbb{E}_{z \sim p(z), s \sim \pi(z)} [\log p (z | s)-\mathbb{E}_{z \sim p(z)} [\log p(z)]] \label{eq:diyan}
\\
&\geq \mathcal{H}[\mathcal{A}|\mathcal{S}, \mathcal{Z}] + \mathbb{E}_{z \sim p(z), s \sim \pi(z)} [\log q_\sigma (z | s)-\log p(z)],
\label{eq:diyan_lower_bound}
\end{align}
The last inequality is due to Jensen’s Inequality. 

\section{Appendix: Algorithm Details}
\subsection{Algorithm}
\begin{algorithm}[H]
\begin{algorithmic}[1]
\STATE \textbf{input}: discriminator parameters $\sigma$, RL parameters $\theta$, code matrix $M_c(z)$, replay buffer, batch size $B$.
\STATE {\bf define} dropout weights $W(t)=(t/T)^2$ for $t=1,\ldots,T$.
\WHILE{not converged}
    \STATE Act (\cref{alg:rollout-collection}). \label{algo:meta:act}
    \STATE sample batch $(s_b, s'_b, t_b, z_b)_{b=1}^B$ from replay buffer. \label{algo:meta:batch}
        \\\textbf{Discriminator predictor}
            \STATE {\bf define} ${y^{\mathrm{ap}}(s_b)} = \tanh(q_\sigma(s_b))$ ($\tanh$ is applied elementwise) for all $b$.
         \label{algo:meta:disc}
        \\\textbf{RL loss}
            \STATE {\bf define} for all $b$, compute $r_b^{\mathrm{apart}} =
            \begin{cases}
                \min_i\{M_c[z_b,i] \, {y^{\mathrm{ap}}(s_b)}\}, & \text{w.p. } W(t_b).\\
                0, & \text{otherwise}.
            \end{cases}$
            \label{algo:meta:reward}
            \STATE {\bf apply} RL algorithm to update $\theta$ w.r.t.\  $r^{\mathrm{apart}}$. \label{algo:meta:update_policy}
        \\\textbf{Discriminator loss}
            \STATE \textbf{set} $\hat{y} =  M_c^\top z$.
            \STATE {\bf define} ${y^\mathrm{ap}}' = \frac{{y^\mathrm{ap}}+1}{2}, \hat{y}'=\frac{\hat{y}+1}{2}$.
            \STATE {\bf define} binary cross entropy loss for the discriminator: $J^\mathrm{ap}(s,z)=-\frac{1}{K}\sum_{i=1}^{K} (y_i^{ap'}\times\log(\hat{y}'_i)+(1-y_i^{ap'})\times\log(1-\hat{y}'_i))$.
            \STATE Update $q_\sigma(s)$ to minimize $J^\mathrm{ap}$ \label{algo:meta:update_disc}
\ENDWHILE
\end{algorithmic}
\caption{Skill discovery with APART}
\label{alg:meta}
\end{algorithm}

\cref{alg:meta} describes the main algorithm loop. Referencing \cref{fig:algo}, line \ref{algo:meta:act} refers to rollout collection (\cref{alg:rollout-collection}), line \ref{algo:meta:disc} refers to computing ${y^{\mathrm{ap}}(s_b)}$ with the discriminator, line \ref{algo:meta:reward} calculates the intrinsic reward (\cref{eq:min_ap}), lines \ref{algo:meta:update_policy}, \ref{algo:meta:update_disc} update policies for the RL and discriminator.

\subsection{Rollout Collection}
\begin{algorithm}[H]
\caption{Rollout collection}
\label{alg:rollout-collection}
\begin{algorithmic}[1]
\STATE \textbf{input}: trajectory length $T$, rollout number to collect $N$, RL parameters $\theta$, replay buffer.
\STATE \textbf{sample} latent variable from prior $z \sim p(\cdot)$.
\STATE \textbf{obtain} RL policy $\pi_{\mathrm{RL}}(\theta, z)$, initial state $s_1$ from environment.
\FOR{$n = 1,\ldots,N$}
    \FOR{$t = 1,\ldots,T$}
      \STATE Sample action $a_t$ from policy $\pi_{\mathrm{RL}}(\theta, z)$ at state $s_t$.
      \STATE Sample next state $s_{t+1}$ from environment after playing $a_t$.
      \STATE Store tuple $(s_t, s_{t+1}, t, z)$ in replay buffer.
    \ENDFOR
\ENDFOR
\end{algorithmic}
\end{algorithm}

\cref{alg:rollout-collection} describes rollout collection, a classic RL setup: the environment outputs a state which is used to choose an action by the RL algorithm. In our case, a latent variable is also sampled from the prior, which is also used by the RL algorithm to choose the action. The transition tuples are stored in a replay buffer to be later used in training.

\subsection{Softmax Temperature Tuning}
\label{appendix:softmax_temp}
\[
\text{softmax}(x) = \frac{e^{\beta x_1}}{\sum_{i=1}^{n} e^{\beta x_i}}
\]

In this formula, \(x\) represents the input vector, \(n\) is the number of elements in the vector, and \(\beta\) is the inverse of the temperature parameter. The softmax function exponentiates each element in the vector with \(\beta\), and then divides it by the sum of all exponentiated elements to obtain the probability distribution. The probabilities sum up to 1, creating a valid probability distribution.

The temperature parameter $\beta\ = 1/T$ controls the smoothness or sharpness of the resulting probability distribution. A lower $\beta$ value makes the distribution more uniform, spreading the probability mass more evenly across all possible outcomes. Conversely, a higher $\beta$ value leads to a more peaked distribution, concentrating the probability mass on a few outcomes with the highest values.

If $\beta$ is very low, approaching zero, the softmax function will tend to assign similar probabilities to all outcomes. This is because extremely high temperatures cause the exponentiated values in the numerator to become more similar, leading to a more uniform distribution.

On the other hand, as $\beta$ approaches infinity, the softmax function becomes more deterministic. The outcome(s) with the highest input value(s) will dominate the probability distribution, receiving almost all of the probability mass, while the probabilities assigned to other outcomes become nearly negligible.

\subsection{Mutual Information Lower Bound}
\label{subsec:lower_bound}
Here, we construct a lower bound on the mutual information with our AP minimum reward.

Similar to other mutual-information algorithms, maximizing the lower bound on the mutual information leads to maximizing the mutual information itself \cite{information_maximization}.
Therefore, we are inspired by MI maximization methods, but do not maximize MI explicitly.

Using \cref{eq:diyan_lower_bound} and \cref{eq:min_ap}, we replace $q(z \mid s)$ in \cref{eq:diyan_lower_bound} with our own discriminator choice, $q'_\sigma(z \mid s)$:
\begin{alignat*}{2}
\mathcal{F}(\theta) 
&\geq \mathcal{H}[\mathcal{A}|\mathcal{S}, \mathcal{Z}] + \mathbb{E}_{z \sim p(z), s \sim \pi(z)} [\log q_\sigma (z | s)-\log p(z)]
\\&\sim \mathcal{H}[\mathcal{A}|\mathcal{S}, \mathcal{Z}] + \mathbb{E}_{z \sim p(z), s \sim \pi(z)} [\log q'_\sigma (z | s)-\log p(z)]
\\&= \mathcal{H}[\mathcal{A}|\mathcal{S}, \mathcal{Z}] + \mathbb{E}_{z \sim p(z), s \sim \pi(z)} [\log e^{\min_{i} \{y'_z(s)[i]\}}-\log p(z)]
\\&= \mathcal{H}[\mathcal{A}|\mathcal{S}, \mathcal{Z}] + \mathbb{E}_{z \sim p(z), s \sim \pi(z)} [\min_{i} \{y'_z(s)[i]\}-\log p(z)]
\end{alignat*}
For reward shaping, we replace $q(z \mid s)$ with $q'_\sigma(z \mid s)$. 
since it gets the semantics of a probability from the training loss, which is the binary cross entropy.

Next, we derive a lower bound on the mutual information using our intrinsic reward. First, we define a softmax on the minimum latent variables as:
\begin{equation}
\label{eq:softmax}
\mathrm{softmax}(q'_\sigma)(z \mid s) = \frac{\exp(q'(z \mid s))}{\sum_{z'} \exp(q'(z'\mid s))}.
\end{equation}

We begin from \cref{eq:diyan}. Our first step is to replace the probability $p(z \mid s)$ with the softmax in \cref{eq:softmax}.
\begin{alignat}{2}
\label{eq:mutual_information}
\mathcal{F}(\theta)  
&\geq \mathcal{H}[\mathcal{A}|\mathcal{S}, \mathcal{Z}] + \mathbb{E}_{z \sim p(z), s \sim \pi(z)} [\log (\mathrm{softmax}(q'_\sigma)(z \mid s)) -\mathbb{E}_{z \sim p(z)} [\log p(z)]]
\\
&= \mathcal{H}[\mathcal{A}|\mathcal{S}, \mathcal{Z}] + \mathbb{E}_{z \sim p(z), s \sim \pi(z)} \Bigg[q'(z \mid s) - \log \bigg(\sum_{z'} e^{q'_\sigma(z' \mid s)} \bigg) -\mathbb{E}_{z \sim p(z)} [\log p(z)] \Bigg]  \label{eq:lower_bound_2}
 \\
&\ge \mathcal{H}[\mathcal{A}|\mathcal{S}, \mathcal{Z}] + \mathbb{E}_{z \sim p(z), s \sim \pi(z)} [q'(z \mid s) - \log(K e) -\mathbb{E}_{z \sim p(z)} [\log p(z)]] \label{eq:lower_bound_3}
\\
&= \mathcal{H}[\mathcal{A}|\mathcal{S}, \mathcal{Z}] + \mathbb{E}_{z \sim p(z), s \sim \pi(z)} [\min_{i} \{M_c[z,i] \, y_i^{ap}[i]\} - \log(K e) -\mathbb{E}_{z \sim p(z)} [\log p(z)]]. \label{eq:lower_bound_4}
\end{alignat}
\cref{eq:mutual_information}: comes from \cref{eq:diyan}, with our own discriminator choice.
\cref{eq:lower_bound_2}: breaking apart nominator and denominator from \cref{eq:softmax}. 
\cref{eq:lower_bound_3}: since $q'_\sigma(z' \mid s)$ is bound by 1, and we sum over all latent variables, K will be an upper bound, where K is the number of latent variables, similarly to eq. (9) in \citep{variational-bounds}.
\cref{eq:lower_bound_4}: using \cref{eq:min_ap}'s full term

$\log p(z)$ is constant in the uniform distribution case and is no longer needed, since it is independent of the skills and discriminator.
its purpose was to normalize the intrinsic reward range to have a zero-mean. In our case, we already achieve a zero mean by using $y_z(s)$ that is in the range $(-1, 1)$, so we can exclude it. 
We are left with $\mathcal{H}[\mathcal{A} \mid \mathcal{S}, \mathcal{Z}] + \mathbb{E}_{z \sim p(z), s \sim \pi(z)} [r_{apart}]$, which is a lower bound on the mutual information and which we optimize with an RL algorithm.

\section{Appendix: Setup}
\subsection{Learning}
\label{subsec_learning}
To run experiments, we chose DQN \citep{DQN} as our RL algorithm. Similar to \citet{DISDAIN}, we act with an epsilon greedy policy which approximates the entropy term.

We use a single fully connected layer for the discriminator backbone and the RL backbone. This is similar to using a lookup table in terms of parameter number needed, but the update method of the parameters is gradient based.
Our observation space is a one hot vector containing only the location of the agent for the discriminator, and the location with the latent variable information for the RL.
Additional implementation choices are explained in \ref{sec:implementation_details}.
Additional choices for parameters and their effect on learning can be found in \cref{subsec:additional_results}.

\subsection{Hyperparameters}
\begin{table}[t]
\caption{Hyperparameters}
\label{table:hyperparameters}
\begin{center}
\begin{tabular}{l|l}
Hyperparameter          & Value       \\ \hline
Torso                   & 1 MLP layer \\
Number of actors        & 1           \\
Batch size              & 640         \\
Skill trajectory length & 40          \\
Number of latents       & 100         \\
Replay buffer size      & 50,000      \\
Optimizer               & ADAM        \\
Learning rate           & 2e-3        \\
RL algorithm            & DQN         \\
$\epsilon$            & 0.001         \\
Discount $\gamma$  & 0.99      
\end{tabular}
\end{center}
\end{table}

\begin{figure}[H]
\begin{center}
\includegraphics[width=1\linewidth]{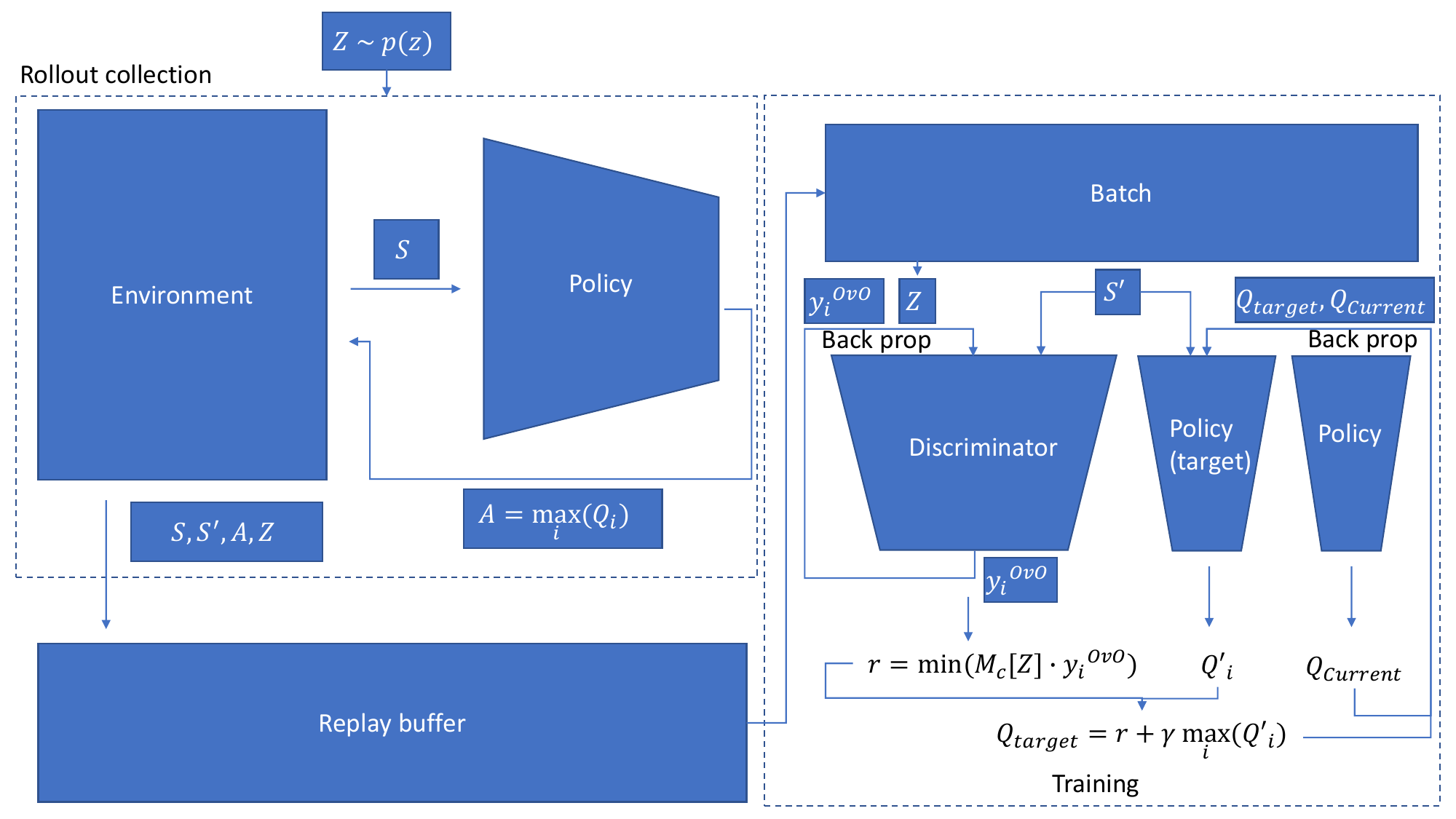}
\end{center}
\caption{Detailed algorithm implementation}
\end{figure}

\subsection{Grid World Environments}
\label{subsec_grid_world_envs}
We use three different grid world environments, each with five actions to choose from: \{Stay, Left, Right, Up, Down\}. 
Although 20 steps are enough to explore almost all states, we show in \cref{fig:diff_env_steps} that learning improves with more environment steps. An agent has more time to explore and is rewarded longer in the final locations.  However, longer episodes are also associated with longer training durations, so we choose an environment step length of 40 steps for $T=40$.

\begin{figure}[H]
\begin{center}
  \subcaptionbox{Four Rooms\label{rooms_example}}
    {\includegraphics[width=0.32\textwidth]{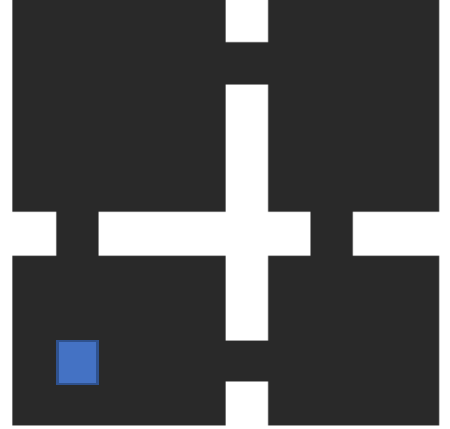}}
  \subcaptionbox{Empty\label{empty_example}}
    {\includegraphics[width=0.32\textwidth]{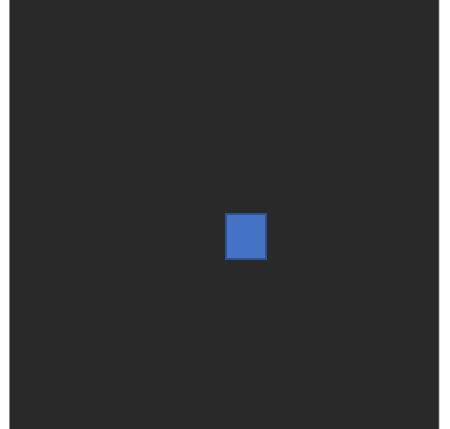}}
  \subcaptionbox{U-maze\label{umaze_example}}
    {\includegraphics[width=0.32\textwidth]{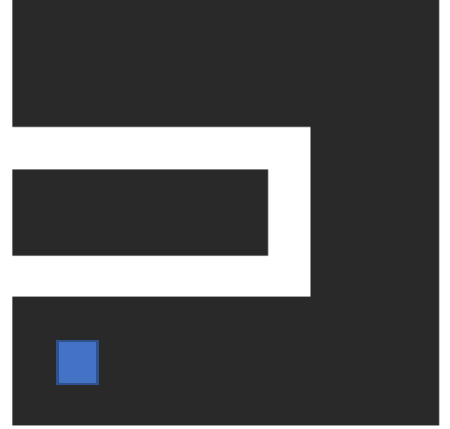}}
  \caption{\textbf{Grid worlds examples, including initial agent location:} (a) Four Rooms (b) Empty (c) U-maze.}
\end{center}
\end{figure}

\textbf{Four rooms.} 10x10 grid separated into four rooms. Each room is connected to the two closest rooms, but the passage is only 1 pixel, forcing many skills to occupy the same state. There are 85 states, the agent's initial position is one step left and one step down from the bottom left corner \citep[\cref{rooms_example}]{sutton1999between, DISDAIN, VIC}.

\textbf{Empty grid.} An empty 10x10 grid. There are 100 states, the agent's initial position is in the middle of the grid (\cref{empty_example}).

\textbf{U-maze.} 10x10 grid, where the whole trajectory is a narrow corridor (3 pixels wide). To reach all states, multiple skills need to advance through the narrow corridor. There are 72 states, the agent's initial position is one step left and one step down from the bottom left corner \citep[\cref{umaze_example}]{direct_then_diffuse}.

\subsection{Metrics}
\label{subsec:metrics}
We measure each algorithm by counting the effective number of skills, i.e. number of occupied last states. 
\begin{align}
  n_{\text{effective}} \triangleq unique(\cup_{i=1}^K S_f(i)) \label{eq:effective_skills}
\end{align}
Where $S_f(i)$ is the last state of skill i, and we count over all latent variables the number of unique last states.
This measure is interpretable as a transformation of mutual information (see \citep{VIC, DIAYN, DISDAIN}) therefore the effective number of skills is an effective and important measure. 

We measure metrics with two configurations: 5M environment steps and 150M environment steps.
On the 5M benchmark, we test for sampling efficiency. In our setting, most of the convergence is in the first million of steps and this is a good indicator to the rest of the training. Furthermore, running a training for hundreds of millions of steps will take the average user weeks and running for 5M steps will usually only take less than one day on a regular computer.
On the 150M benchmark, we test for full skill acquisition, meaning whether $n_{\text{effective}} = n_{\text{available}}$, where $n_{\text{available}}$
 represents all of the available states (effectively, grid size without walls)

\section{Appendix: Implementation Choices} \label{sec:implementation_details}
\label{subsec:additional_results}
In this section, we review implementation choices and their effect on learning. This is done by fixing all parameters except for one, and choosing 3-4 options for the tested parameter. If the value is numerical, desirably, we'll choose at least one option smaller than our chosen hyperparameter and one larger. If the value is categorical, desirably we'll want to show a few different choices and explain their similar and different characteristics. All graphs in this section are on Rooms grid environment.
For example, we will discuss what batch size is best out of four options, learning rates, number of environment steps, etc.

\begin{figure}[H]
\begin{center}
\includegraphics[width=0.85\linewidth]{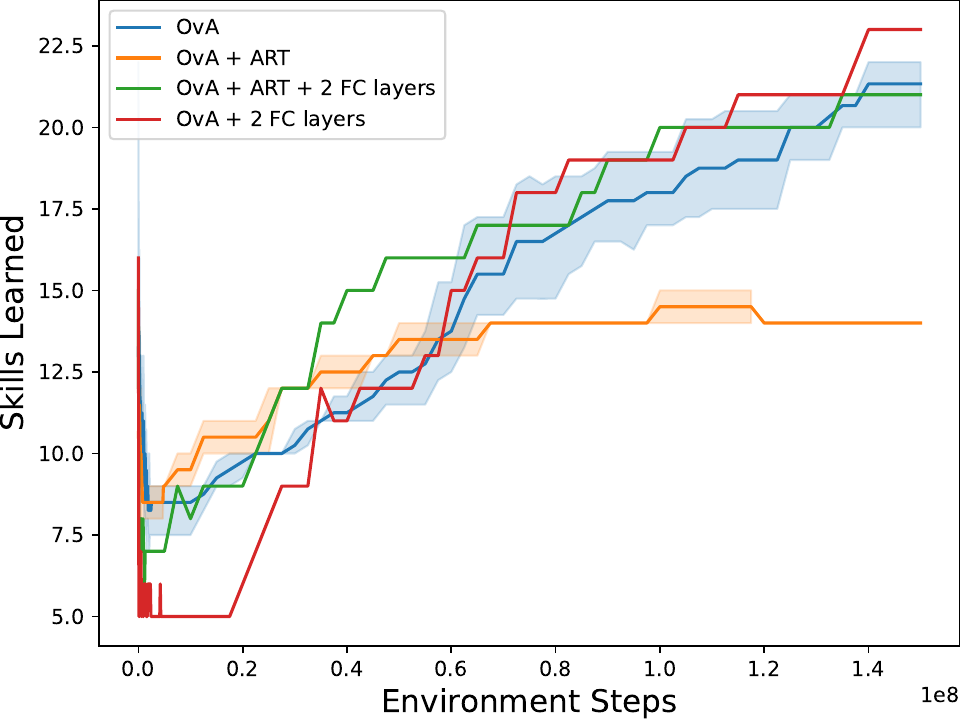}
\end{center}
\caption{One vs All with two FC layers vs one. ART stands for ascending reward and dropout.}
\label{fig:ova_2layers}
\end{figure}

\paragraph{Number of NN layer weights.}
Our setup uses only one fully connected layer as our backbone. Therefore, number of parameters used (weights) is the number of inputs times the number of outputs (plus bias terms).
The number of network outputs is $L=K(K-1)/2$ for AP but only $K$ for OvA, resulting in a significantly larger number of weights for AP.
We therefore want to test the effect of the amount of weights between the two methods. 
For this, we added another fully connected layer to the OvA method with 2300 weights (to get a similar number of parameters between methods). We calculate number of weights as follows: rooms environment has 85 available states and we use 100 latent variables. Therefore, for OvA, we have 85 inputs and 100 outputs. The regular setup will result in about 85x100=8500 parameters, while after adding a hidden layer we will have 85x2300+2300x100=425,500 parameters. AP has same number of inputs (85) but uses $K(K-1)/2$ outputs and therefore uses about 85x100x99/2=420,750 weights. 
Furthermore, adding another layer means adding a non-linearity which is helpful for learning. 
In \cref{fig:ova_2layers} we show that adding a hidden layer of size 2300 does not lead to a significant change in learning, indicating that it is not the number of weights constraining the learning for OvA.

\begin{figure}[H]
\begin{center}
\includegraphics[width=0.85\linewidth]{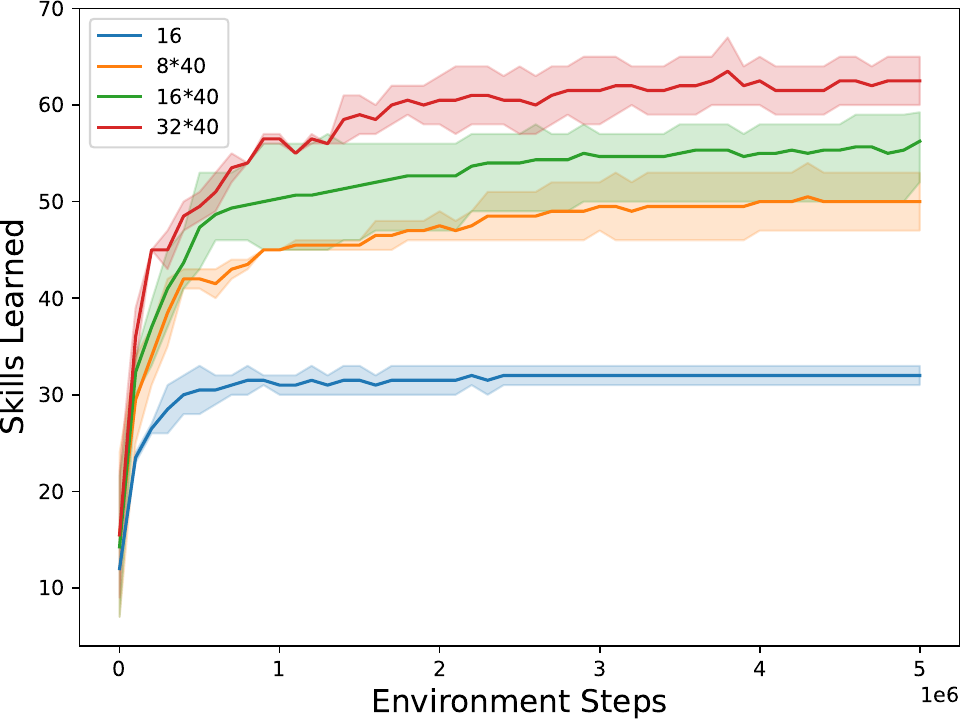}
\end{center}
\caption{different batch sizes}
\label{fig:batch_size}
\end{figure}

\paragraph{Batch size.}
The number of samples taken from the replay buffer in each iteration of APART (Line~\ref{algo:meta:batch} in \cref{alg:meta}) has an effect on both smoothness of the training and compute resources. While the batch size is bounded by the size of the GPU memory (if using a GPU), a larger batch size also increases training time and can slow down convergence. 
A smaller batch size, on the other hand, can lead to unstable learning. 
We test different values of this hyperparameter in multiples of environment steps. 
Although we sample randomly from the replay buffer, not necessarily taking entire rollouts, we still want to keep an explainable multiple of the environment steps for the batch size, meaning we test our batch sizes as multiples of number of chosen environment steps (40).
In \cref{fig:batch_size}, we see the effect of different choices, including 16 (choice from \citep{DISDAIN}), 320 (8 times rollout length), 640 (16 times rollout length, this is our chosen parameters, in order to be able to run multiple runs simultaneously) and 1280 (32 times rollout length) on the number of learned skills.

\begin{figure}[H]
\begin{center}
\includegraphics[width=0.85\linewidth]{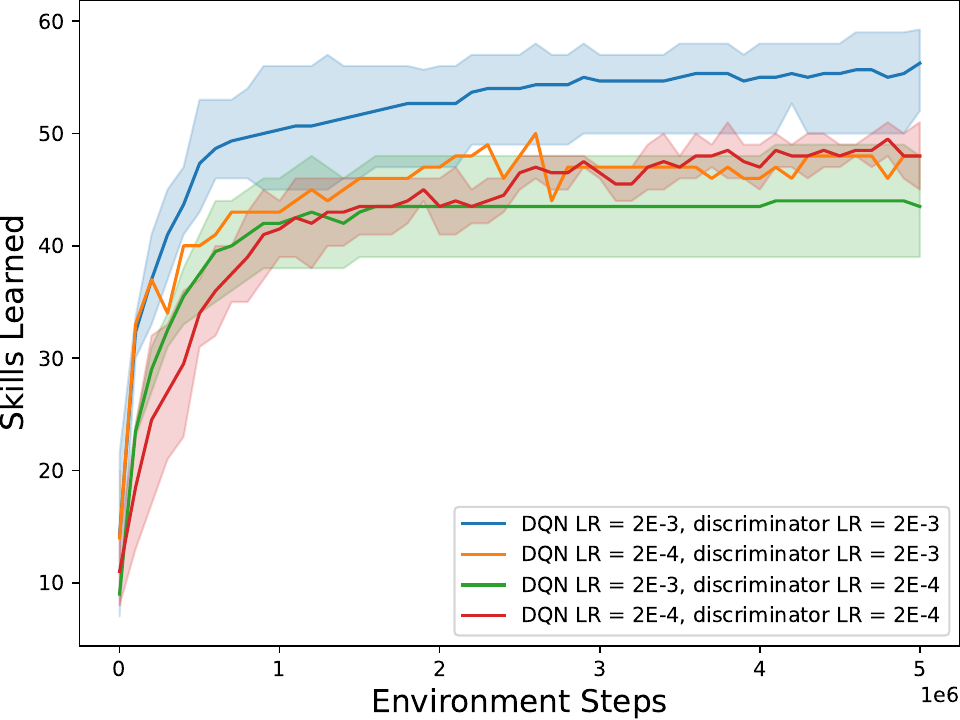}
\end{center}
\caption{different learning rates}
\label{fig:learning_rates}
\end{figure}

\paragraph{Learning rate.}
Our algorithm trains two NN's simultaneously. 
One is trained using the DQN algorithm and the other implements the discriminator. 
The learning rate of these two NNs are tightly coupled since the reward for the RL is derived from the discriminator outputs, and the state from the RL algorithm is sampled from the policy chosen by the DQN. 
If the discriminator were to learn too slowly, the reward will be far from the true estimation. 
If the RL algorithm were to learn too slowly, its policy won't be meaningful for the discriminator to learn from. 
In \cref{fig:learning_rates} we can see the effect of different learning rates on learning, our chosen learning rate is $2\mathrm{E}{-3}$ for discriminator and $2\mathrm{E}{-3}$ for the RL algorithm.

\begin{figure}[H]
\begin{center}
\includegraphics[width=0.85\linewidth]{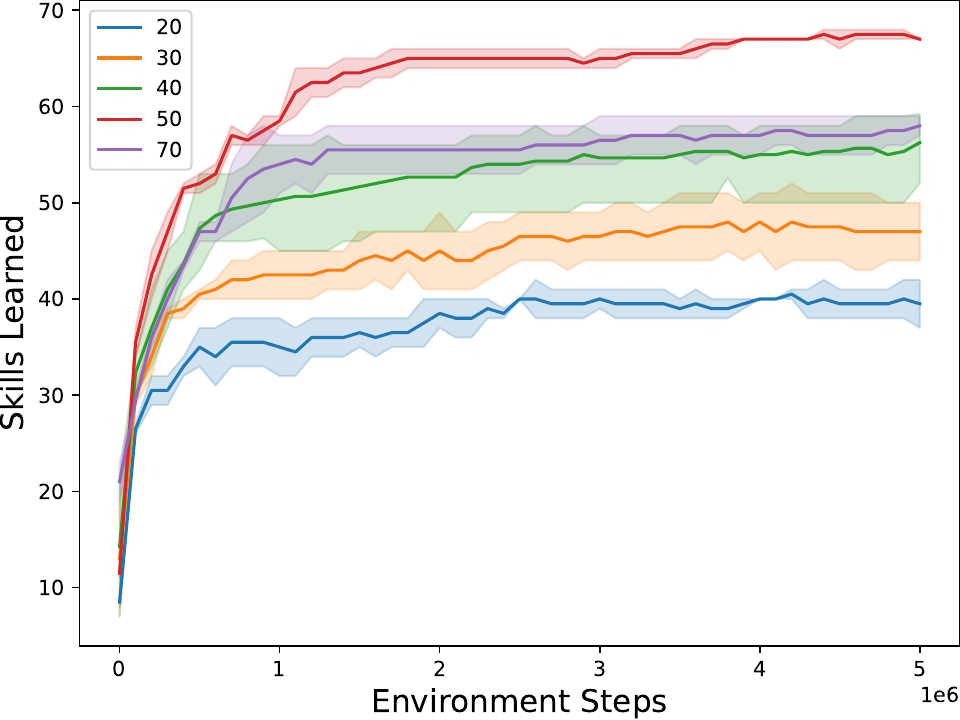}
\end{center}
\caption{different env steps}
\label{fig:diff_env_steps}
\end{figure}

\paragraph{Environment step length.} 
Our environments don't have a ``natural'' end to them (collisions for example), thus the only termination point is determined by the environment step length. 
Although 20 environment steps are enough to reach all possible states, additional allowed steps mean more possible paths to the same state. 
Moreover, staying on the last state for a longer period of time enables more collected transitions in which the state and skill are coupled (the same latent variable leads to the same state), therefore strengthening the link between them. 
On the other hand, we do need to terminate the episode at some point in order to keep collecting more episodes.
In \cref{fig:diff_env_steps}, we can see a few different choices for environment step lengths. 
\cite{DISDAIN} use 20 environment steps, yet we observe that this is not an optimal choice. 
Our choice of 40 environment steps is also not an optimal choice, and in hindsight we might have chosen 50 environment steps for our experiments, but because of limited time and resources we used 40.

\begin{figure}[H]
\begin{center}
\includegraphics[width=0.85\linewidth]{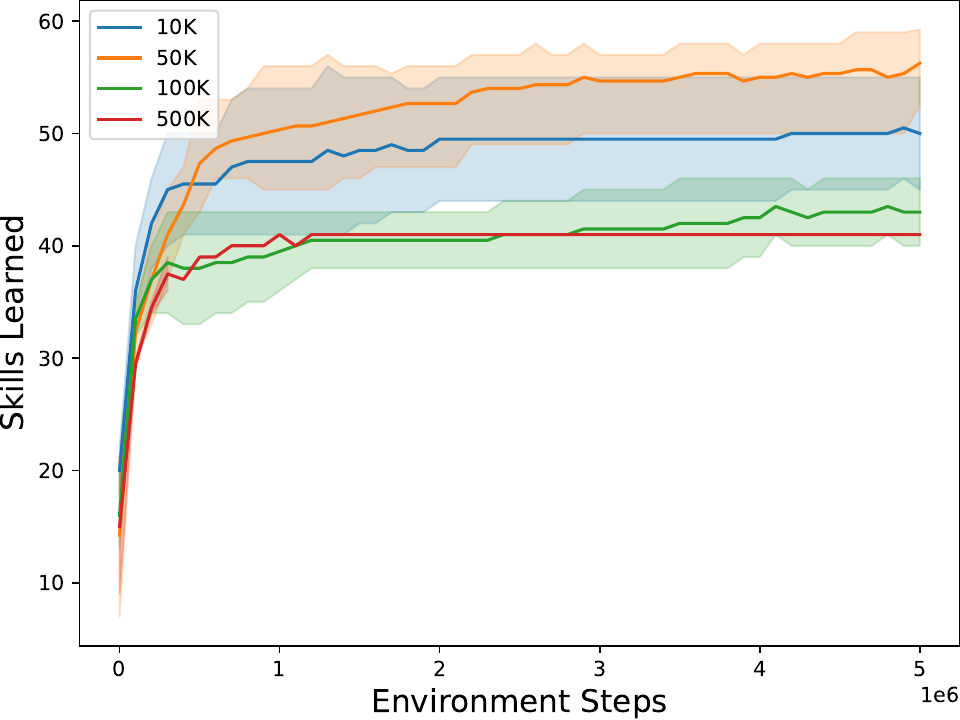}
\end{center}
\caption{different buffer sizes}
\label{fig:diff_buff_size}
\end{figure}

\paragraph{Replay buffer size.}
The replay buffer holds transitions $(s, s', a, z)$ from rollouts, and is sampled randomly during training. 
When collecting new transitions, the oldest transitions are replaced by the new sampled rollouts. 
The larger the buffer, the larger the transition history is and the more offline the learning is, meaning the learning is further away from the current policy, since older samples that use older policies are used. 
On the other hand, a small buffer can result in instability during training since the effect of each collected rollout will be larger and won't be smoothed out by the effect of the older transitions \citep{replay_buffer}.
In \cref{fig:diff_buff_size}, we can see the effect of the buffer size in our settings. 
We can see that the gap between a 10K buffer and a 50K buffer is small whereas the gap between 50K and 100K is significant. Our choice for this hyperparameter is 50K as it shows the best empirical results.

\begin{figure}[H]
\begin{center}
\includegraphics[width=0.85\linewidth]{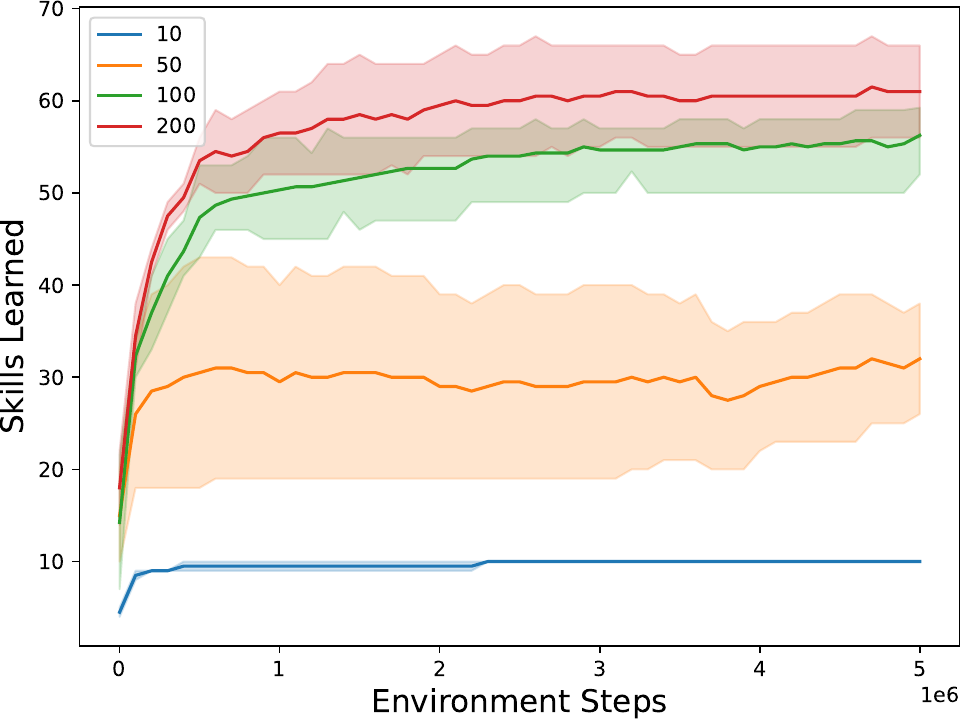}
\end{center}
\caption{Number of learned skills vs different number of latent variables}
\label{fig:diff_latent_num}
\end{figure}

\paragraph{Number of latents.} 
We experiment with different number of latents and see their effect on learning. 
For each rollout, we select a prior out of a distribution size, which is the number of latents. For example, if we use 100 latents, our prior is chosen as a number between 0 and 99.
Since our metric for diversity is the number of unique last states \cref{eq:effective_skills}, and there is a constant and known number of last states per environment, we would like to pick number of latents which is close to the number of possible states. In \cref{fig:diff_latent_num}, we explore different choices, including our choice which is 100 latents, but also a much smaller choice of 10, a choice of 50 and a larger choice of 200.

\begin{figure}[H]
\begin{center}
    \includegraphics[width=0.85\linewidth]{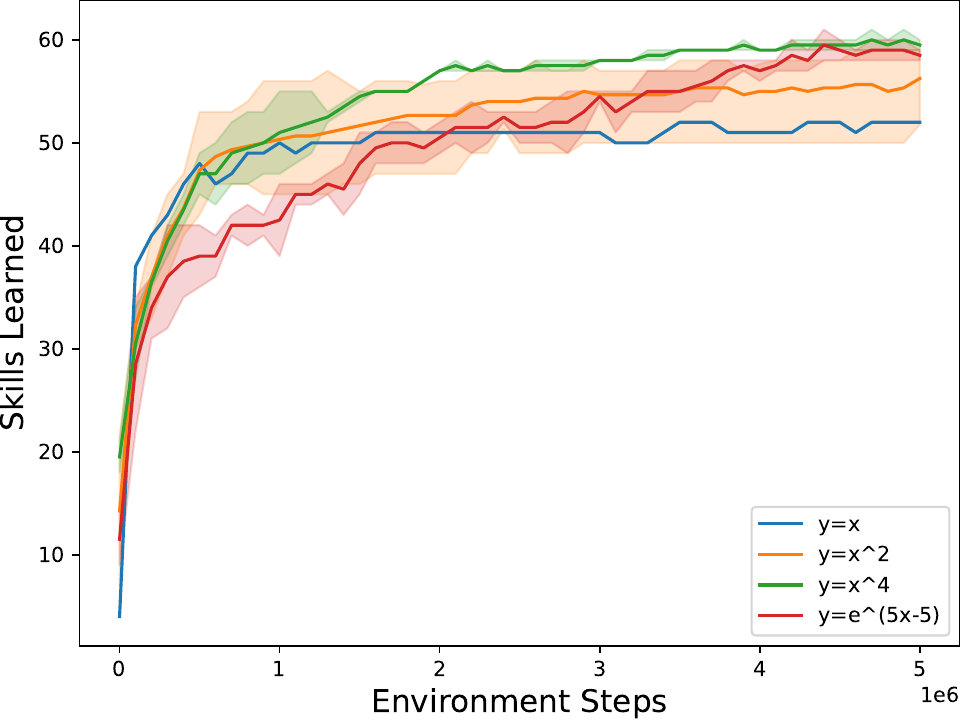}
\end{center}
\caption{\textbf{Different weight functions over 5 million environment steps in rooms.} Blue: linear function, orange: power of 4, green: $\exp(5t-5)$, red: squared function (APART).}
\label{fig:diff_weight_options}
\end{figure}

\paragraph{Ascending reward function.} 
The ascending function \cref{eq:dropout} in our algorithm is used to add weights to emphasize the reward as the step number increases.
We want the final steps to have a larger weight than the initial steps, since in the initial steps the skills are tightly packed and harder to discriminate (since all agents have the same initial position). 
Therefore, it is important to choose an ascending function. 
In \cref{fig:diff_weight_options}, we pick a few options (for the positive $x$ values corresponding to step number): $W(t)=t, W(t)=t^2, W(t)=t^4, W(t)=e^{5t-5}$. We show that the exact choice of an ascending function for the weights is not as important, since their performance is similar.

\begin{figure}[H]
\begin{center}
\includegraphics[width=0.85\linewidth]{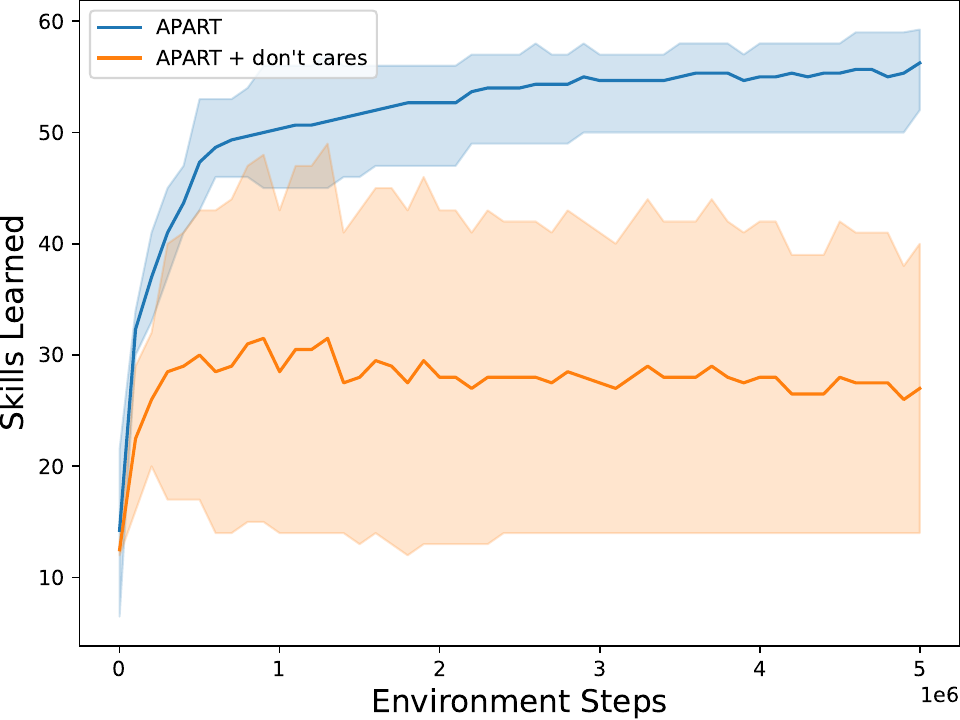}
\end{center}
\caption{Adding "don't cares" to discriminator}
\label{fig:ap_with_dont_cares}
\end{figure}

\paragraph{Using ``don't cares'' in AP discriminator.} 
When using AP classification, \cref{sec:ova_and_ap}, we use pair comparisons between each two classes $i,j$, resulting in $K(K-1)/2$ comparisons.
The ``don't cares'' are the pair comparisons from the AP classification that are indifferent to the result. 
For example, if the true class is 0 and the pair comparison is between 1 and 2, the prediction can ignore the classifier that classifies between 1 and 2. 
One option (used in \cite{ONEVSONE}) is to use this information in the discriminator's loss, with a value of 0 (or 0.5 when scaling outputs to the range (0, 1)). 
In \cref{fig:ap_with_dont_cares} we revisited this choice and compare it with only backpropagating on meaningful pair comparisons (masking out all "don't care" comparisons). Meaning, discriminator loss is only given $K-1$ meaningful comparisons, instead of $K(K-1)/2$ comparisons which hold some meaningful comparisons but also "don't cares".
We found that in our setting, it is better to mask out the ``don't cares''.
Another advantage of masking don't cares is backpropagating over only $K-1$ values instead of $L=K(K-1)/2$ values.

\begin{figure}[H]
\begin{center}
\includegraphics[width=0.85\linewidth]{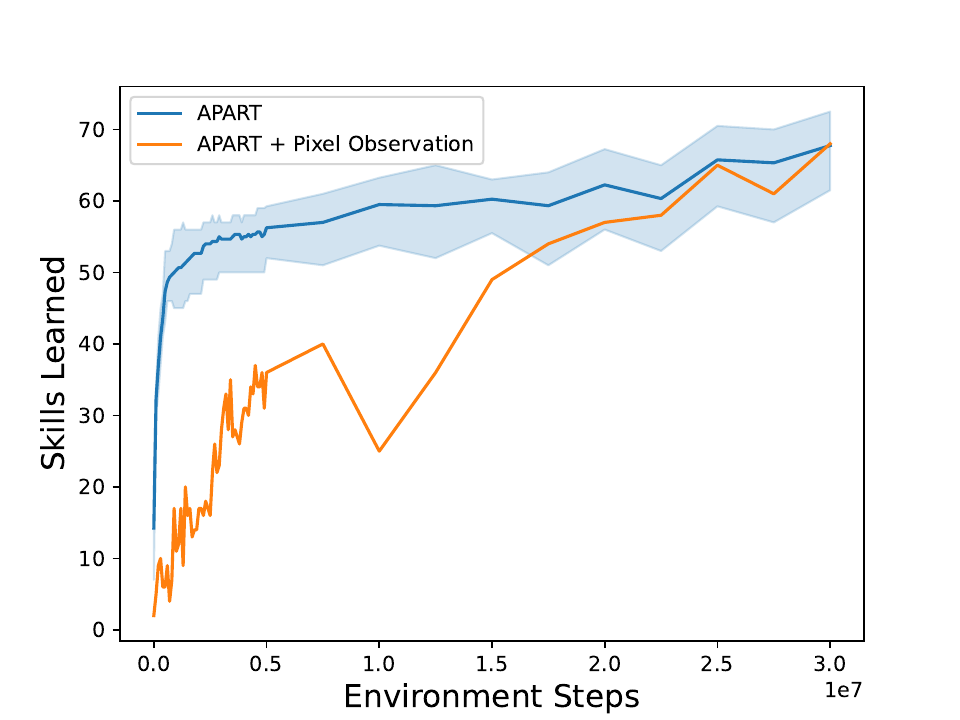}
\end{center}
\caption{Pixel observation space}
\label{fig:pixel_observation}
\end{figure}

\paragraph{Using different observation spaces.} 
As explained in \cref{sec:rl}, the observation space is an implementation choice. We initially chose to use a one-hot vector observation \cref{subsec_learning} but other options are also available.
In order to further investigate our method, and in preparation to run on more complex environments (Atari \citep{atari} for example), we want to see its performance with a pixel-based observation space, meaning the observation space is an image containing all grid information, as a human player would observe when playing the game.
To accommodate a pixel-based observation space, we also change our backbones to CNNs \citep{alexnet}, as is the regular choice for image-based observation spaces, and adjust our learning rates accordingly. We do these changes for both the discriminator and the RL.
The backbone used has 3 CNN layers with feature sizes of (32, 64, 128), followed by an FC layer. 
We also use padding of 2, considering we don't want to lose any of our input pixels, and stride of 1. 
We don't use any pooling for the same reason. 
The learning rates we use are $1\mathrm{E}{-5}$ for the RL algorithm and $3\mathrm{E}{-4}$ for the discriminator. 
The rest of the parameters remain unchanged. 
We can see in \cref{fig:pixel_observation} that we are able to achieve good learning with this observation space as well, paving the way to experimenting on more complex environments.

\section{Appendix: Additional Results} 
\label{sec:additional-results}
\paragraph{Reward Comparison.}
Reward comparison of all methods found in \cref{fig:reward_comparison}, with the addition of VIC. It is evident that VIC stands out due to its significantly larger scale and distinctiveness compared to other methods.
\begin{figure}[H]
\begin{center}
    {\includegraphics[width=0.85\textwidth]{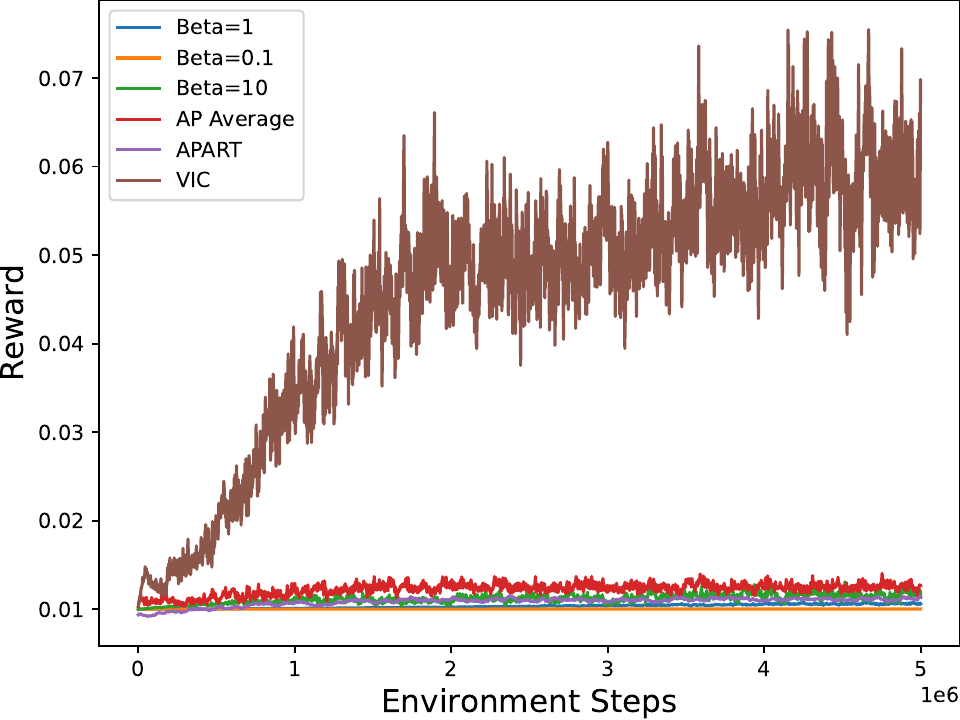}}
    \hfill
    \hfill
  \caption{Comparison of the learned rewards from six different methods: VIC, APART and AP Average, Tuned VIC with: $\beta=10$, $\beta=1$ and $\beta=0.1$. Rewards are smoothed with $\sigma=5$.}
\label{fig:reward_comparison_appendix}
\end{center}
\end{figure}

\paragraph{Learned Skills}

Up until now, we only talked about number of learned skills without providing additional information on the skills learned. 
In \cref{fig:plot_all_skills}, we show all learned skills during the training process, on different environments and for different methods. 
Skills are sorted by last state visited and are numbered by their latent ID. 
It is interesting to note that not all states are reached in the fastest way. Another observation is that since our latent number is larger than the possible states number, we will always have a few overlapping skills. 
We can also see the skills evolving through time in different algorithms.

\begin{figure}[H]
\begin{center}
  \subcaptionbox{Rooms, APART\label{Rooms, APART}}
    {\includegraphics[width=1\linewidth]{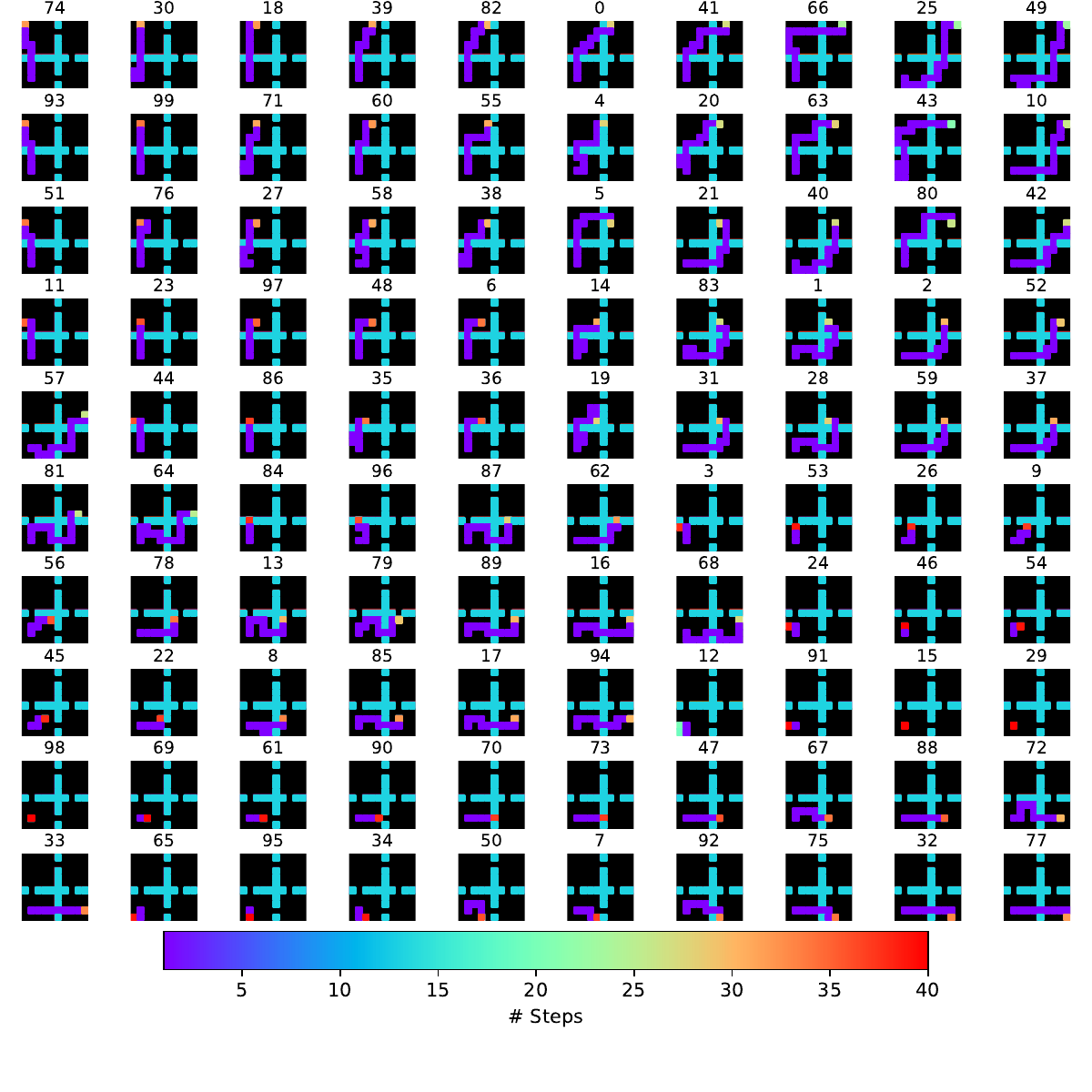}}
\end{center}
\end{figure}

\begin{figure}[H]
\begin{center}
  \ContinuedFloat
  \subcaptionbox{Rooms, VIC\label{Rooms, VIC}}
    {\includegraphics[width=1\linewidth]{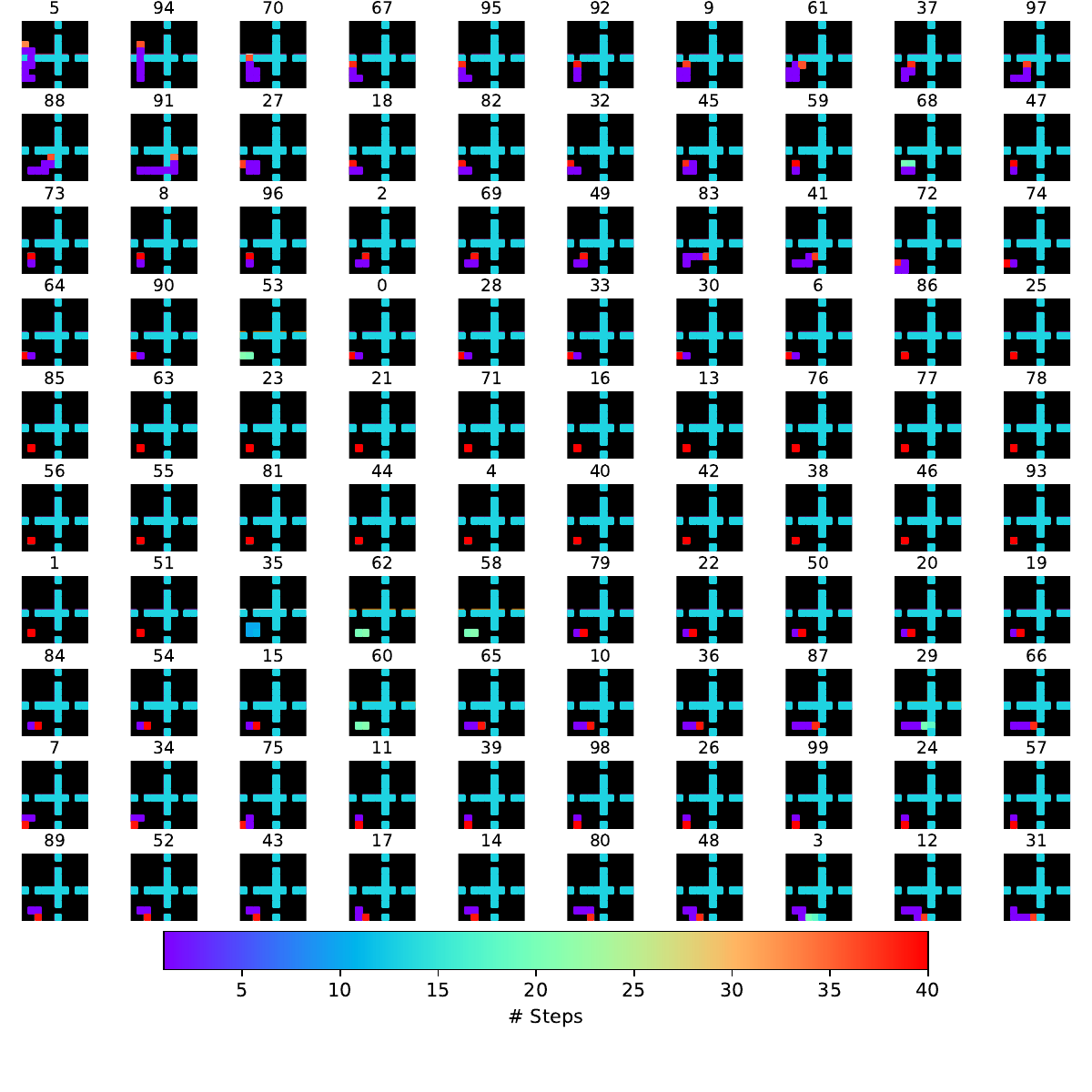}}
\end{center}
\end{figure}

\begin{figure}[H]
\begin{center}
  \ContinuedFloat
  \subcaptionbox{Rooms, DIAYN\label{Rooms, DIAYN}}
    {\includegraphics[width=1\linewidth]{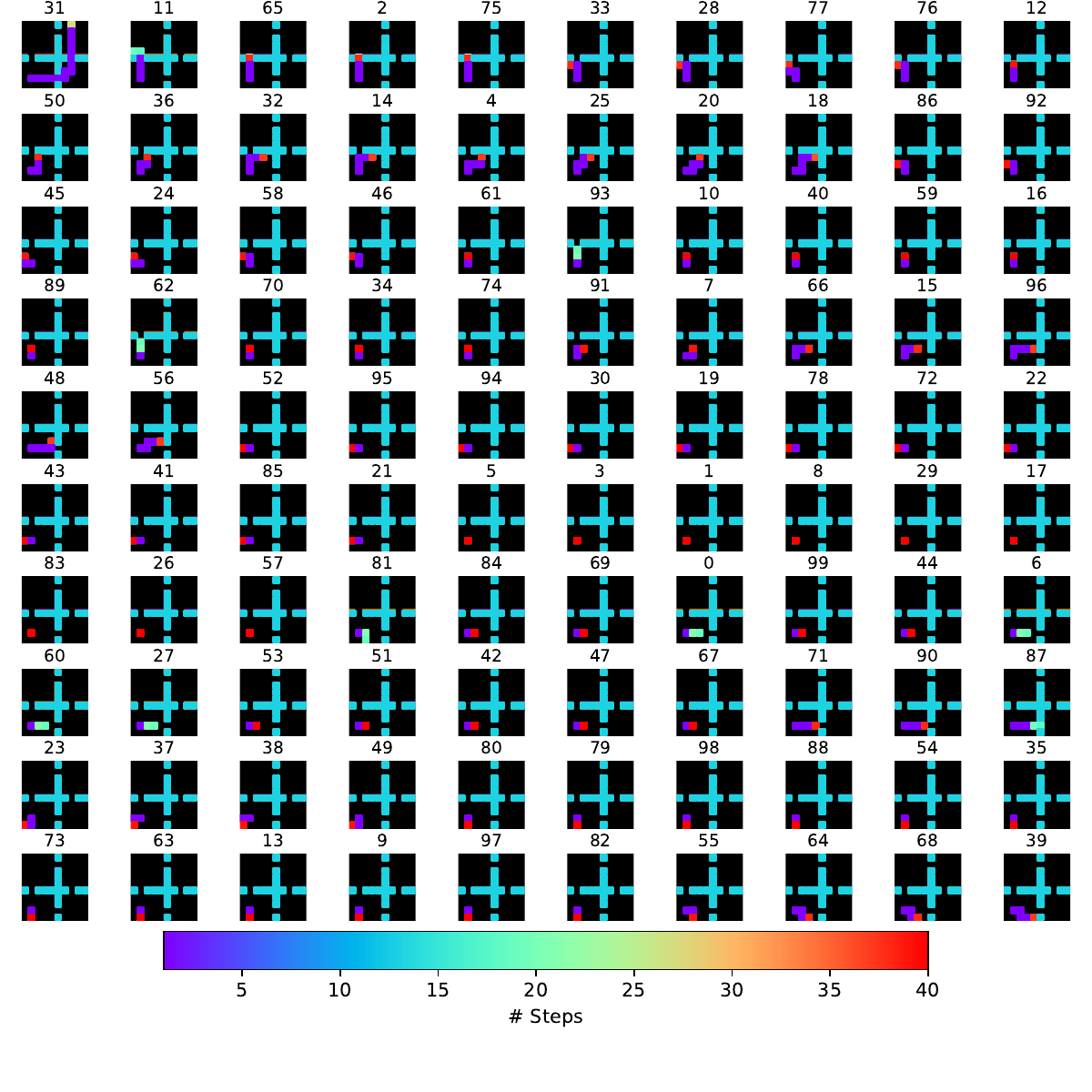}}
\end{center}
\end{figure}

\begin{figure}[H]
\begin{center}
  \ContinuedFloat
  \subcaptionbox{Empty, DIAYN\label{Empty, DIAYN}}
    {\includegraphics[width=1\linewidth]{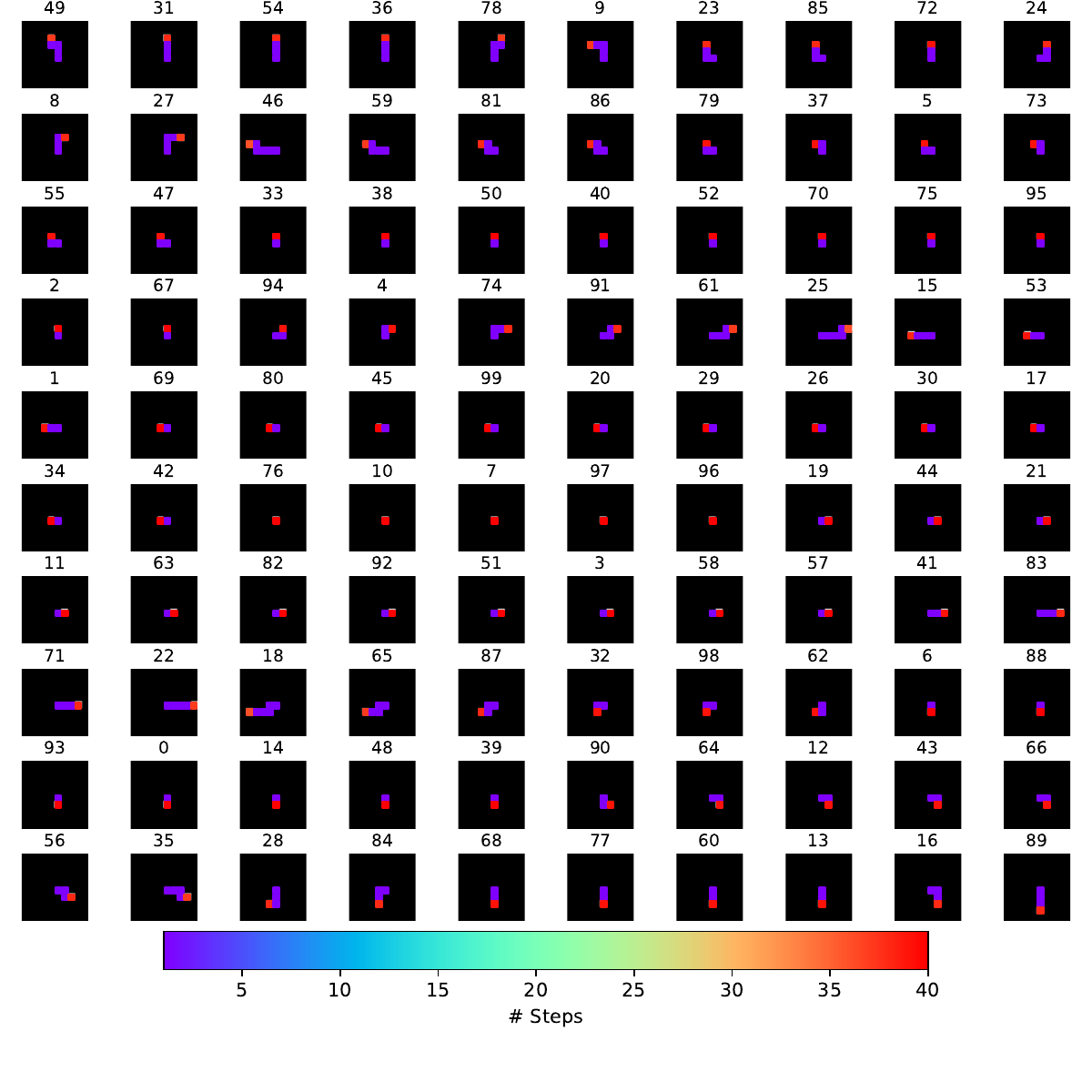}}
\end{center}
\end{figure}

\begin{figure}[H]
\begin{center}
  \ContinuedFloat
  \subcaptionbox{Empty, VIC\label{Empty, VIC}}
    {\includegraphics[width=1\linewidth]{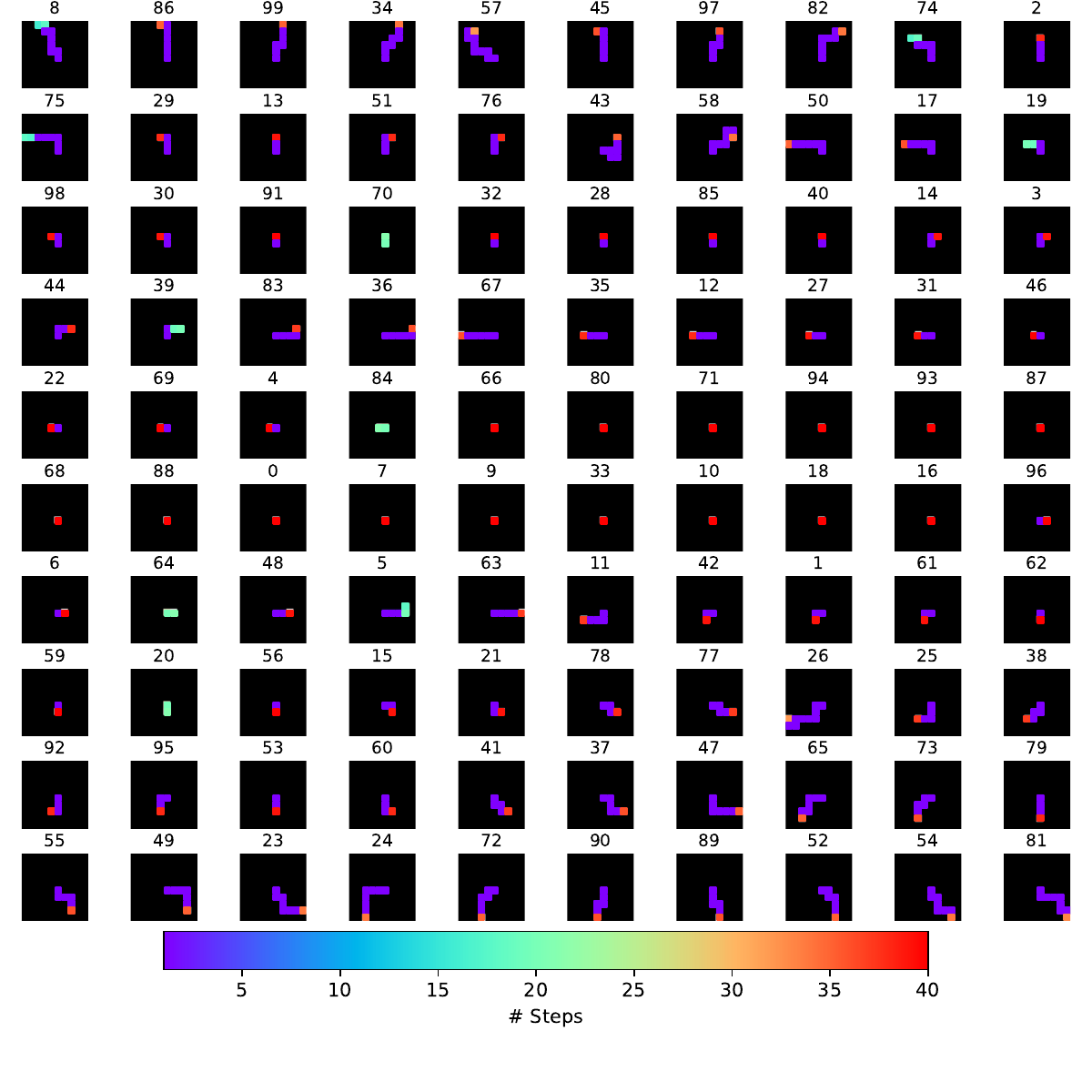}}
\end{center}
\end{figure}

\begin{figure}[H]
\begin{center}
  \ContinuedFloat
  \subcaptionbox{Empty, APART\label{Empty, APART}}
    {\includegraphics[width=1\linewidth]{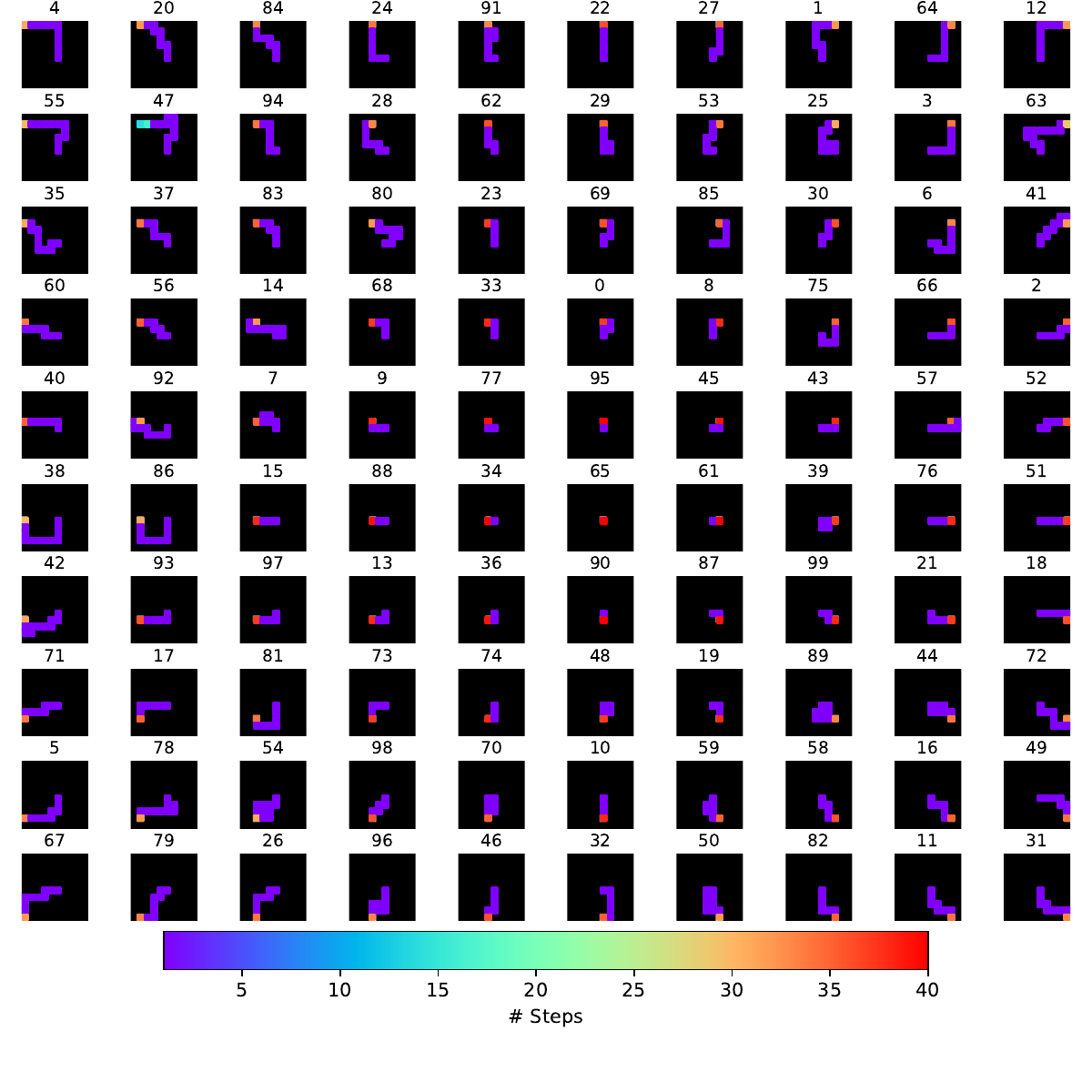}}
\label{fig:plot_all_skills}
\caption{\textbf{All skills after convergence of algorithm}, skills are ordered by last state.}
\end{center}
\end{figure}

\end{document}